\documentclass{article} % For LaTeX2e
\usepackage{iclr2027_conference,times}

\usepackage{amsmath,amsfonts,bm}

\def\eqref#1{equation~\ref{#1}}
\def\1{\bm{1}}

\DeclareMathAlphabet{\mathsfit}{\encodingdefault}{\sfdefault}{m}{sl}
\SetMathAlphabet{\mathsfit}{bold}{\encodingdefault}{\sfdefault}{bx}{n}

\usepackage[utf8]{inputenc}
\usepackage[T1]{fontenc}
\usepackage{xcolor}
\definecolor{cbgreen}{RGB}{34,136,51}
\definecolor{cbblue}{RGB}{0, 119, 187}
\definecolor{cbred}{RGB}{204, 51, 17}
\definecolor{richlilac}{rgb}{0.71, 0.4, 0.82}
\definecolor{mydarkblue}{rgb}{0,0.08,0.45}
\definecolor{circuitred}{RGB}{255, 220, 220}    % light red (object circuit, as in the figures)
\definecolor{circuitblue}{RGB}{210, 225, 245}   % light blue (relational circuit, as in the figures)
\newcommand{\circbox}[2]{{\setlength{\fboxsep}{0.8pt}\colorbox{#1}{\strut #2}}}
\usepackage{url}
\usepackage{booktabs}
\usepackage{amsfonts}
\usepackage{amsmath}
\usepackage{amssymb}
\usepackage{amsthm}
\theoremstyle{definition}
\newtheorem{definition}{Definition}
\usepackage{nicefrac}
\usepackage{microtype}
\usepackage{graphicx}
\usepackage{subcaption}
\usepackage{enumitem}
\usepackage{wrapfig}
\usepackage{float}
\usepackage{listings}
\usepackage{placeins}
\usepackage[colorlinks=true,citecolor=cbgreen,linkcolor=magenta,urlcolor=mydarkblue]{hyperref}
\usepackage{cleveref}

\title{Matching Object or Relation? \\
Tracing Abstract Reasoning Inside VLMs}

\author{Gouki Minegishi$^{1}$ \\[0.25em]
\textbf{Hiroki Furuta}$^{2}$\thanks{Work done as an advisory role only.} \quad
\textbf{Takeshi Kojima$^{1}$} \quad
\textbf{Yusuke Iwasawa$^{1}$} \quad
\textbf{Yutaka Matsuo$^{1}$} \\[0.5em]
$^{1}$The University of Tokyo \quad
$^{2}$Google DeepMind \\[0.25em]
\texttt{minegishi@weblab.t.u-tokyo.ac.jp}}

\iclrfinalcopy
\begin{document}

\maketitle
% Override ICLR camera-ready header for arXiv preprint
\lhead{Preprint}

\begin{abstract}
Vision Language Models (VLMs) excel on visual benchmarks but fail systematically on tasks requiring abstract reasoning. Existing benchmarks document this failure but cannot say \emph{why} it happens or which cognitive capability is missing. We close this gap by adopting the Relational Match-to-Sample (RMTS) paradigm from comparative and developmental psychology and pairing it with a mechanistic analysis of the model's internals. On a parametrically controlled stimulus set evaluated across frontier API models (GPT, Claude, Gemini) and three open-source families (Qwen3.5, Gemma-4, InternVL3), we identify four levers that shift VLMs toward the relational match---capability tier, model scale, the number of objects per scene, and the absence of per-object stimulus noise---together producing a developmental-like trajectory that mirrors the human \emph{relational shift}. Opening up the model, a per-layer representational similarity analysis and a causal mediation analysis reveal that VLM abstract reasoning is implemented by two competing circuits: an early circuit that organises images by their surface object features, and a late circuit that organises them by their abstract relation. 
Extending the analysis to ARC-AGI-1, we find that ablating the relational heads identified on RMTS degrades performance more than ablating random heads, indicating that the relational circuit is recruited beyond our controlled stimuli. 
We hope this mechanism-level view serves as a step toward understanding how abstract reasoning is implemented in VLMs.
\end{abstract}

%=============================================================
\section{Introduction}
\label{sec:intro}
%=============================================================

Vision Language Models (VLMs) have advanced rapidly on visual tasks~\citep{yue2024mmmu, yue2024mmmu_pro, li2025survey, zhang2024multimodalchainofthoughtreasoninglanguage, li2025visuallargelanguagemodels}, yet they continue to fail systematically on a kind of \emph{abstract reasoning}~\citep{mitchell2023comparing, campbell2024understanding, yiu2024kiva}. Existing benchmarks document this failure but conflate it with object recognition, attention, and configuration tracking; the resulting picture is descriptive rather than mechanistic, leaving open both \emph{why} VLMs underperform and which cognitive capability they lack relative to mature human intelligence. A diagnostic that ties behaviour to internal mechanism has been missing.

\begin{figure}[t]
\centering
\includegraphics[width=0.84\linewidth]{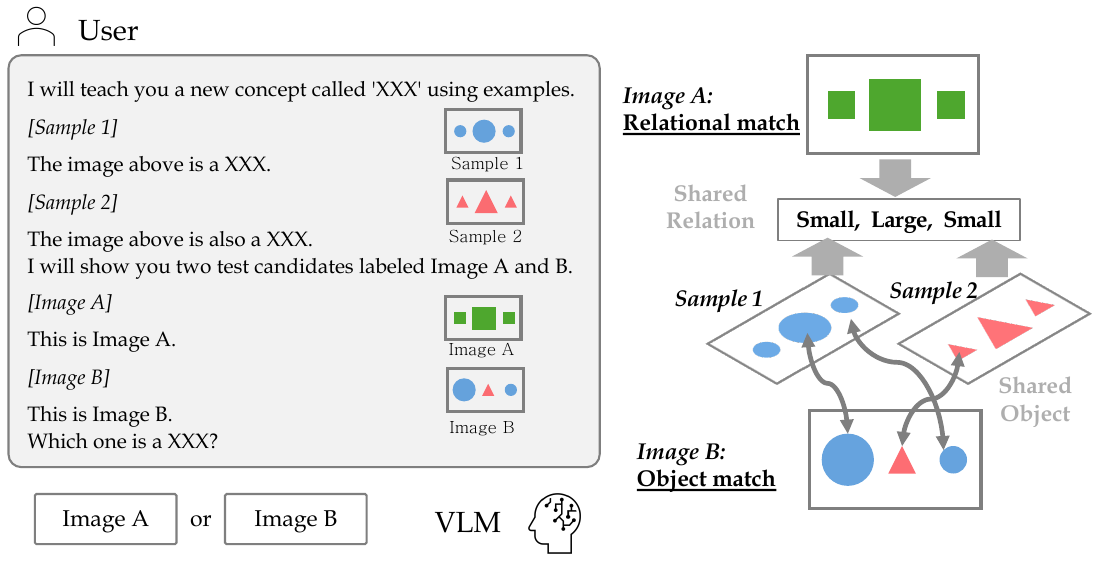}
\caption{\textbf{The Relational Match-to-Sample (RMTS) task as we present it to a VLM.}
\textbf{Left:} the prompt as the model sees it. The VLM is taught a novel nonsense label \emph{XXX} from two example images (Sample 1 and Sample 2), and is then asked which of two test candidates---Image A or Image B---is also a \emph{XXX}. \textbf{Right:} the two competing solutions the model can latch onto. Image A (top) preserves the \texttt{small--large--small} relation with brand-new objects (green squares); the label \emph{XXX} therefore generalises to A only if the model has bound it to the \emph{shared abstract relation} between the samples---the \emph{relational match}. Image B (bottom) reuses the very objects that appeared in the samples (a blue circle and red triangles) but breaks the size pattern; the label generalises to B only if the model has bound it to \emph{shared surface features}---the \emph{object match}. The two readings are mutually exclusive by construction: a relational reasoner picks A, while a system that keys on surface object similarity picks B.}
\label{fig:task_example}
\vspace{-5mm}
\end{figure}

To close this gap, we adopt the \textbf{Relational Match-to-Sample (RMTS)} paradigm from comparative and developmental psychology~\citep{LanguageHelpsChildren, rmst, analogyandabstraction}, illustrated in \autoref{fig:task_example}. Two sample images are introduced with a novel nonsense label (\emph{e.g.}, ``these are both \emph{XXX}''), and the model is asked which of two test images is also a \emph{XXX}. The \emph{relational match} preserves the abstract pattern shared by the samples---here, a \texttt{small--large--small} size pattern---with entirely new objects; the \emph{object match} reuses the samples' own objects (a blue circle and a red triangle) but breaks the pattern. The question is not which option is objectively correct, but which kind of generalisation the system reaches for. Adult humans gradually shift toward the relational match---a developmental trajectory known as the \emph{relational shift}~\citep{Gentner2011StructureMA, LanguageHelpsChildren, goswami1991}---while young children and non-human apes begin near the object-match end~\citep{ApeChild, manandbeast}. 
RMTS thus offers a calibrated yardstick for relational abstraction in any cognitive system, with established human-developmental and cross-species reference points.

We bring this paradigm to VLMs. 
To eliminate confounds from naturalistic scene content, we design a formally controlled stimulus set whose scenes consist of geometric objects parameterised by \texttt{(size, color, shape)}, and evaluate frontier API models (GPT, Claude, Gemini) alongside three open-source families (Qwen3.5, Gemma-4, InternVL3). On this task we make two contributions.

\textbf{(1) A behavioural map of the object--relational axis in VLMs. (\Cref{sec:behavior})} We identify four levers that systematically shift VLMs toward the relational match: capability tier (a $58$\,pp gap inside a single API family between Claude-Opus-4.6 and Claude-Haiku-4.5), within-family scale, the number of objects per scene $n$, and the absence of per-object stimulus noise. Chain-of-thought prompting modulates the first two by $+20$ to $+57$\,pp on non-saturated models. The resulting picture is a developmental-like trajectory across model families and conditions, mirroring the human relational shift.

\textbf{(2) A mechanistic account of how the choice is made. (\Cref{sec:internal})} Behaviour alone does not explain \emph{why} models choose as they do. Using a per-layer representational similarity analysis~\citep[RSA;][]{rsa}, we show that early layers organise the four images in line with the object-match solution, while late layers swing toward the relational-match solution, with the crossover arriving earlier and sharper in larger and more relationally-biased models. A causal mediation analysis~\citep[CMA;][]{meng2022, wang2022} isolates a single dominant relational head and a single dominant object head per checkpoint, with logit-margin swings of order one unit when patched. Together, RSA and CMA reveal that VLM abstract reasoning is implemented by two competing circuits: an early circuit that organises images by their surface object features, and a late circuit that organises them by their abstract relation. When the model answers directly, its choice reflects which of the two circuits dominates at the output; CoT, by contrast, reaches the relational answer through a route that does not depend on the relational heads.

We hope that bringing controlled, mechanism-aware experimental designs of this kind to VLMs can serve as a step toward a research programme on abstract reasoning that goes beyond aggregate benchmark scores and engages directly with how that capability is implemented inside the model.
\clearpage
%=============================================================
\section{The Relational Match-to-Sample Setup}
\label{sec:task}
%=============================================================
\begin{wrapfigure}{r}{0.36\textwidth}
    \vspace{-15pt}  
    \centering
    \includegraphics[width=0.35\textwidth]{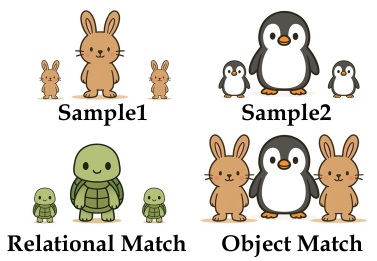}
    \caption{
    Typical RMTS stimuli in cognitive science.}
    \label{fig:example}
    \vspace{-10pt}  
\end{wrapfigure}

In cognitive science, RMTS is typically conducted with naturalistic scenes containing multiple objects such as animals~\citep{analogyandabstraction}. For instance, as illustrated in \autoref{fig:example}, stimuli are constructed by treating entities such as rabbits or penguins as the units over which relational and object matches are defined. 
In contrast, to probe how VLMs perform abstract reasoning under minimal confounds, we adopt a tightly controlled stimulus design in which the attributes of each object are explicitly parameterised. 
This design follows a line of work on controlled task construction in deep-learning vision research~\citep{assouel2025visual, NEURIPS2024_99e6bcf4, NEURIPS2023_9d0f188c}. Concrete examples of our stimuli are shown in \autoref{fig:task_example} and, more comprehensively, in \autoref{fig:stimuli_canonical} in Appendix~\ref{app:task_example}.

\subsection{Trial structure: samples, relational match, object match}
\label{sec:task:trial}

We make the structure of an RMTS trial precise. Each scene is built from objects parameterised by three attributes drawn from finite alphabets:
\begin{equation*}
    \mathcal{C} = \{\text{red}, \text{blue}, \text{green}\}\,, \qquad
    \mathcal{S} = \{\text{small}, \text{medium}, \text{large}\}\,, \qquad
    \mathcal{H} = \{\text{circle}, \text{triangle}, \text{square}\}\,,
\end{equation*}
the colour, size, and shape sets respectively. An object is a triple $o = (c, s, h)\in\mathcal{C}\times\mathcal{S}\times\mathcal{H}$.

\begin{definition}[Samples]
A \emph{sample} of size $n$ is an ordered tuple of $n$ objects that share a common shape\footnote{Fixing the shape within a sample preserves the setting illustrated in \autoref{fig:example}: it lets shape act as a perceptual grouping cue that signals which objects belong to the same scene, while keeping the relation carried entirely by colour and size. Allowing shape to vary within a sample would make the task substantially harder, as the model would also need to identify the grouping itself.}:
\begin{equation*}
    \mathbf{x} = (o_1, \ldots, o_n)\,, \qquad o_i = (c_i, s_i, h)\;\;\forall i\,.
\end{equation*}
The within-scene patterns are the equivalence classes of the colour and size sequences:
$P_\mathcal{C}(\mathbf{x}) = \mathrm{pat}(c_1, \ldots, c_n)$ and $P_\mathcal{S}(\mathbf{x}) = \mathrm{pat}(s_1, \ldots, s_n)$, where for $n=3$ we use the standard alphabet $\mathrm{pat}\in\{\textsc{AAA}, \textsc{AAB}, \textsc{ABA}, \textsc{ABB}, \textsc{ABC}\}$ (e.g.\ $\mathrm{pat}(\text{small}, \text{large}, \text{small}) = \textsc{ABA}$). The \emph{relation} of a scene is the pair $R(\mathbf{x}) = \bigl(P_\mathcal{C}(\mathbf{x}),\, P_\mathcal{S}(\mathbf{x})\bigr)$.
\end{definition}

A trial is a tuple $T = \bigl(\mathbf{x}^{(1)}, \ldots, \mathbf{x}^{(K)}; \mathbf{y}_{\mathrm{rel}}, \mathbf{y}_{\mathrm{obj}};\, \ell\bigr)$ consisting of $K\!\geq\!2$ \emph{sample scenes} $\mathbf{x}^{(k)}$ that share the same relation $R^* = R(\mathbf{x}^{(k)})$ but use distinct shapes and distinct attribute-value assignments, two \emph{test scenes} $\mathbf{y}_{\mathrm{rel}}$ and $\mathbf{y}_{\mathrm{obj}}$, and a label $\ell$ (a novel three-letter nonce word) attached to each sample. The model is asked which of $\mathbf{y}_{\mathrm{rel}}, \mathbf{y}_{\mathrm{obj}}$ is also an $\ell$.

\begin{definition}[Relational match]
\label{def:relmatch}
A scene $\mathbf{y}_{\mathrm{rel}}$ is a \emph{relational match} for samples sharing relation $R^*$ if (i) its relation matches: $R(\mathbf{y}_{\mathrm{rel}}) = R^*$; (ii) its shape is not used by any sample, $h(\mathbf{y}_{\mathrm{rel}}) \notin \{h(\mathbf{x}^{(k)})\}_k$; and (iii) its concrete colour and size values are disjoint from those used by the samples (when the alphabet permits).
\end{definition}

\begin{definition}[Object match]
\label{def:objmatch}
A scene $\mathbf{y}_{\mathrm{obj}}$ is an \emph{object match} if (i) its relation \emph{differs} from the samples', $R(\mathbf{y}_{\mathrm{obj}}) \neq R^*$; (ii) every object in $\mathbf{y}_{\mathrm{obj}}$ is drawn from the union of objects appearing in the samples, $\{o : o \in \mathbf{y}_{\mathrm{obj}}\} \subseteq \bigcup_k \{o : o \in \mathbf{x}^{(k)}\}$; and (iii) the shapes used in $\mathbf{y}_{\mathrm{obj}}$ are mixed across $\geq 2$ of the sample shapes.
\end{definition}

The two definitions are mutually antagonistic by construction. \Cref{def:relmatch} guarantees that picking $\mathbf{y}_{\mathrm{rel}}$ requires inferring the abstract relation from the samples and applying it to genuinely new objects (new shape, new colour/size values); \Cref{def:objmatch} guarantees that picking $\mathbf{y}_{\mathrm{obj}}$ requires only matching surface features back to the samples (because every object in $\mathbf{y}_{\mathrm{obj}}$ literally appears in some sample). 
A system that keys on object similarity therefore deterministically picks $\mathbf{y}_{\mathrm{obj}}$; a system that keys on the relation $R^*$ deterministically picks $\mathbf{y}_{\mathrm{rel}}$. 
Following the cognitive-science convention \citep{analogyandabstraction, LanguageHelpsChildren} we report the rate at which the model picks $\mathbf{y}_{\mathrm{rel}}$ (the \emph{relational rate}, Rel\%) and treat the contrast as a \emph{bias} rather than an accuracy measurement. 

\subsection{Prompt and evaluation protocol}
\label{sec:task:eval}

Each trial is presented in a single turn containing all four images followed by a short instruction (see concrete prompt in Appendix~\ref{app:prompts}). 
The label is a novel three-letter nonce word drawn from a fixed pool to avoid priming any pre-existing concept; the question is phrased neutrally (``Which one is a XXX?'') and constrained with ``Final answer: Image~'' so that the next token is forced to be either \textsc{A} or \textsc{B}. We sweep three open-source families covering 0.8B--38B parameters---Qwen3.5~\citep{team2026qwen3} (0.8B/2B/4B/9B/27B), Gemma-4~\citep{gemma4_2026} (E2B/E4B/26B/31B), and InternVL3 \citep{internvl3} (1B/2B/8B/14B/38B)---together with  frontier VLM models from OpenAI (GPT-5.4, GPT-5.4-mini)~\citep{openai_gpt54thinking_2026}, Anthropic~\citep{anthropic_opus46_2026} (Claude-Opus-4.6, Claude-Haiku-4.5), and Google~\citep{gemini31pro_2026} (Gemini-3.1-Pro). For open-source models we read the raw logits of the tokens \textsc{A} and \textsc{B} directly from the next-token distribution, avoiding free-text parsing; for closed-source API models we parse the generated response. 
% In either case the prediction is the higher-scoring option.

Two sources of bias must be removed for a relational rate to be interpretable.
First, the choice of letter labels (A/B) and of nonce word can each induce a constant preference; we cycle the nonce word across a fixed pool of five three-letter strings---XXX, YYY, ZZZ, QQQ, and WWW---assigned uniformly at random per trial, and average the relational rate over labels, integrating out any nonce-specific preference. The pool is restricted to repeated-character strings so that no element carries lexical content that could prime a real concept. 
Second, the model's prior over the answer position can favour one letter---a hard ``A-bias'' is observable in some open-source checkpoints. 
To remove this, every trial is presented in both orderings (relational match as \textsc{A} and as \textsc{B}) and the relational rate is averaged over the two; any constant letter preference therefore cancels. Unless stated otherwise, the relational rate (Rel\%) is reported over all trials per model.
By default, we use the $n=3, k=2$ configuration shown in \autoref{fig:task_example}, which yields $729$ distinct trials per model; together with the two label orderings used to remove A-bias, this gives $729 \times 2$ evaluations per model. Full configuration and model details are deferred to \Cref{app:stimuli_stats,app:models}.

%=============================================================
\section{Behaviour: When VLMs Pick the Relational Match}
\label{sec:behavior}
%=============================================================

% \emph{Every VLM we tested can pick the relational match given sufficient scale, prompting, or stimulus saliency.} 
We characterise relational match bias along four levers---(1) capability tier with chain-of-thought (CoT) in \Cref{sec:behavior:frontier}, (2) model-size scaling in \Cref{sec:behavior:scale}, (3) number of objects $n$ in \Cref{sec:behavior:n}, and (4) stimulus noise in \Cref{sec:behavior:noise}. The levers are instances of two factors: the \emph{capability of the model} (tier and scale, with CoT as an inference-time counterpart), the analogue of the maturation that drives the human relational shift; and the \emph{strength of the relational evidence in the stimulus} ($n$ and noise).\footnote{More objects make a shared property overwhelming and each object less individuating; per-object noise degrades exactly the colour/size evidence that defines the relation.}
% ---and report an attribute breakdown in \Cref{app:attr_breakdown} (\autoref{tab:attr_breakdown}). 
% Headline numbers are computed on the canonical 729-trial set with position debiasing (\Cref{sec:task:eval}); full per-model breakdowns appear in \Cref{app:cot_full}.

\begin{figure}[t]
\centering
\begin{subfigure}[t]{0.32\linewidth}
\includegraphics[width=\linewidth]{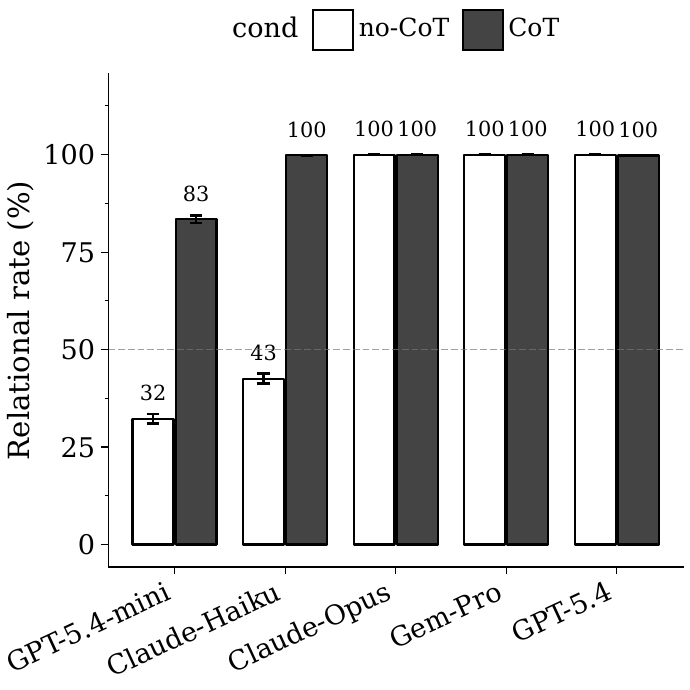}
\caption{Frontier VLMs.}
\label{fig:behavior:frontier_cot}
\end{subfigure}\hfill
\begin{subfigure}[t]{0.32\linewidth}
\includegraphics[width=\linewidth]{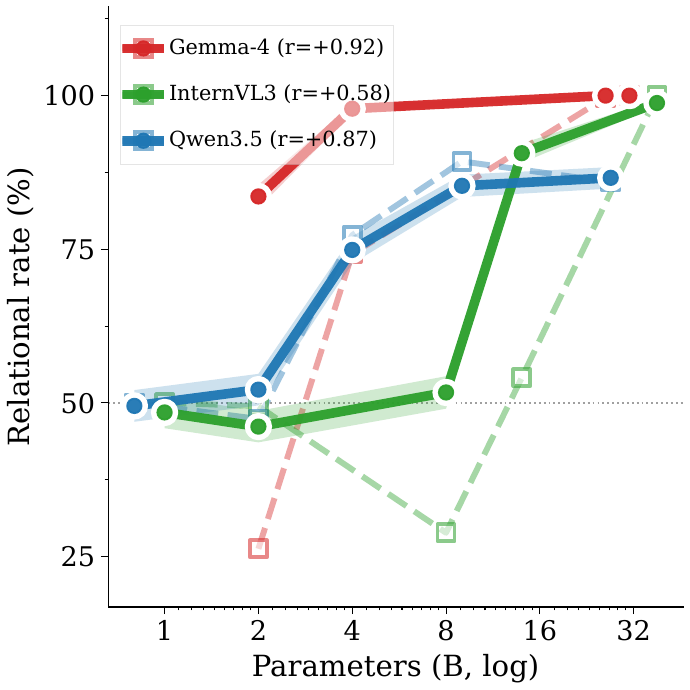}
\caption{Model scale.}
\label{fig:behavior:scale}
\end{subfigure}\hfill
\begin{subfigure}[t]{0.32\linewidth}
\includegraphics[width=\linewidth]{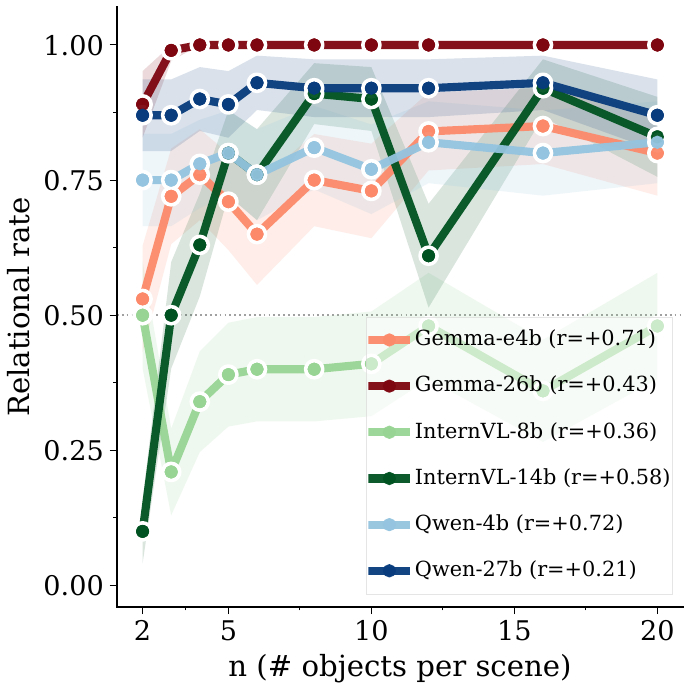}
\caption{Number of objects $n$.}
\label{fig:behavior:n}
\end{subfigure}
\caption{\textbf{Capability tier, model scale, and number of objects each promote the relational match.}
(a) Capability tier drives a $58$\,pp spread within Claude (Opus $100$\% vs.\ Haiku $42.5$\%); CoT lifts every non-saturated model ($+51$ to $+57$\,pp). (b) Within each open-source family, larger checkpoints pick the relational match more reliably (solid lines, filled circles: CoT; dashed lines, open squares: no-CoT). (c) Raising the number of objects per scene from $n{=}2$ to $n{=}20$ shifts the relational rate upward.}
\label{fig:behavior:row1}
\vspace{-5mm}
\end{figure}

\subsection{Frontier VLMs show a wide object--relational spread}
\label{sec:behavior:frontier}

We begin by measuring RMTS performance across five frontier VLMs---GPT-5.4-mini, Claude-Haiku-4.5, Claude-Opus-4.6, Gemini-3.1-Pro, and GPT-5.4---as shown in \autoref{fig:behavior:frontier_cot}. Strikingly, even on this set of frontier models, the no-CoT condition reveals a pronounced spread in relational rate, ranging from 32\% for GPT-5.4-mini to 100\% for GPT-5.4 (a 68-pp gap). Adding chain-of-thought prompting consistently shifts models toward the relational match: the two weakest models, GPT-5.4-mini and Claude-Haiku-4.5, gain 51-pp (32 $\to$ 83) and 57-pp (43 $\to$ 100) respectively, while the three already-saturated models remain at 100\%.\footnote{The set is designed as an instrument for analysing \emph{how} a model answers rather than as a benchmark; for closed-source models it does act as one and is at ceiling, but a harder relation (numerical equality) brings them off it (GPT-5.4: $82.5$\%; \Cref{app:numeq}).} 
Following \citet{kojima2022large}, we elicit CoT with a ``Let's think step by step'' prompt; the exact prompt template is given in Appendix~\ref{app:prompts}. Inspecting the generated reasoning traces, we find that the models explicitly extract the abstract relation shared between the two samples and apply it to the test scenes---example traces are provided in Appendix~\ref{app:cot_outputs}.

\subsection{Model scaling drives the relational match}
\label{sec:behavior:scale}

Within each open-source family, larger models pick the relational match more reliably (\autoref{fig:behavior:scale}; per-model rates in \autoref{tab:cot_full}). Gemma-4 rises monotonically from $26.3$\% (E2B) to $100$\% (31B), and Qwen3.5 from $50.0$\% (0.8B) to $89.3$\% (9B), with a $-3.4$\,pp dip at 27B that is at the noise floor. InternVL3 reaches $100$\% at 38B; its apparent dip at 8B ($28.9$\%) arises because 1B and 2B sit at chance, committing to neither candidate, whereas 8B commits to the object match before scale flips it.\footnote{The rank correlation between parameter count and relational rate is positive within every family (Spearman $\rho=0.80$, $1.00$, $0.60$ for Qwen3.5, Gemma-4, InternVL3; $\rho=0.78$ over all fourteen checkpoints), and the scale effect also holds for numerical equality (\Cref{app:numeq}).} The no-CoT rate correlates with $\log$-model size in Gemma-4 ($r=+0.92$), Qwen3.5 ($+0.80$) and, more weakly, InternVL3 ($+0.58$), and CoT trajectories sit above the no-CoT ones in every family. The aggregate picture is a developmental-like trajectory mirroring the human relational shift \citep{LanguageHelpsChildren, analogyandabstraction}: surface-dominant at small scale, relation-dominant at large scale.

\subsection{More objects per scene promote the relational match}
\label{sec:behavior:n}

Raising the number of objects per scene from $n{=}2$ to $n{=}20$ shifts the relational rate upward in the object-biased middle tiers of each family (\autoref{fig:behavior:n}, per-model Pearson $r(n,\mathrm{Rel})$ in the legend; example stimuli in \autoref{fig:stimuli_n}). Gemma-4-E4B climbs from $0.53$ to $0.80$ ($r\!=\!+0.71$), InternVL3-14B from $0.10$ to $0.83$ ($r\!=\!+0.58$) and Qwen3.5-4B from $0.75$ to $0.82$ ($r\!=\!+0.72$); InternVL3-8B shows a weaker trend ($r\!=\!+0.36$), and the near-ceiling Gemma-4-26B and Qwen3.5-27B show small positive correlations around saturation. As $n$ grows, the abstract relation becomes visually salient, and the surface heuristic that succeeds at $n{=}2$ (``the candidate that reuses the samples' objects looks more familiar'') no longer suffices. The same trend holds for GPT-5.4-mini, the only non-saturated frontier model in our suite (\autoref{fig:behavior:n_api}).

\subsection{Stimulus noise weakens the relational match}
\label{sec:behavior:noise}

The controlled stimuli used so far are sharp idealisations: every object has a categorical colour (red/blue/green) and a categorical size (small/medium/large), rendered at fixed pixel values. Real-world visual tasks (\emph{e.g.}, natural-image analogues of RMTS) are messier---two ``red'' objects rarely have identical pixel values, and two ``small'' objects rarely have identical pixel sizes. To probe how robust the relational behaviour observed above is to this kind of low-level perturbation, we add per-object Gaussian noise to colour and size while leaving the categorical identity of each object unchanged. Concretely, for each object $o$ in a scene we apply
\begin{equation}
    \mathrm{rgb}_o \leftarrow \mathrm{rgb}_o + \mathcal{N}(\mathbf{0}, \sigma_c^2 \mathbf{I}_3)
    \quad\text{and}\quad
    \mathrm{size}_o \leftarrow \mathrm{size}_o + \mathcal{N}(0, \sigma_s^2),
\end{equation}
clipped to valid pixel and size ranges. We sweep $\sigma_c \!\in\! \{0,5,10,20,40,80\}$ (RGB units) and $\sigma_s \!\in\! \{0,1,2,4,8\}$ (px) with three independent noise seeds per cell. Examples of the resulting stimuli across the noise sweep are shown in \autoref{fig:noise_examples}.

\begin{figure}[t]
\centering
\includegraphics[width=0.86\linewidth]{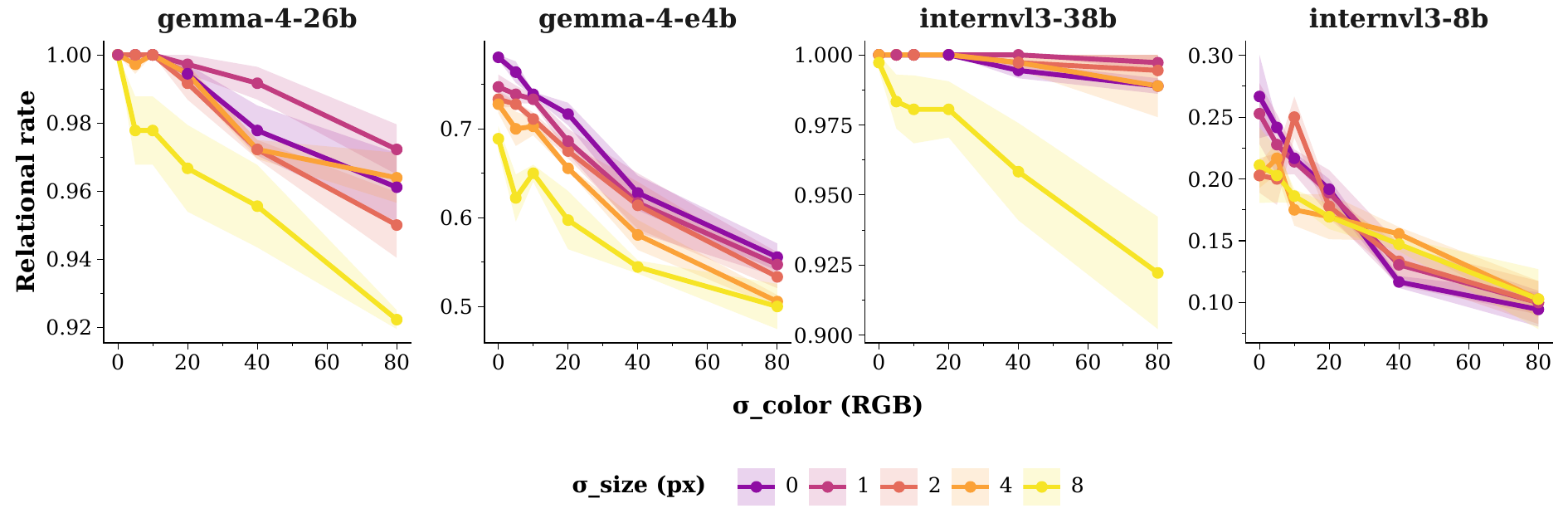}
\caption{\textbf{Per-object stimulus noise reduces the relational rate.} 
Relational rate as a function of colour-noise scale $\sigma_c$ (RGB units; $x$-axis) for four open-source checkpoints, with size-noise scale $\sigma_s$ (px) encoded by line colour. Both noise channels degrade the relational rate, and within each family the smaller checkpoint degrades more steeply. Shaded bands: $\pm 1$ s.e.\ over three noise seeds per cell.}
\label{fig:behavior:noise}
\vspace{-5mm}
\end{figure}

\autoref{fig:behavior:noise} shows the results for four open-source checkpoints (Gemma-4-26B/E4B, InternVL3-38B/8B). Two patterns are immediately visible.
First, both noise channels degrade the relational rate: every panel slopes downward as $\sigma_c$ grows, and within each panel the curves separate by $\sigma_s$ (with the largest $\sigma_s = 8$ uniformly the lowest). Second, smaller models within each family are markedly more sensitive to noise. Gemma-4-26B drops only from $\sim\!100$\% (clean) to $\sim\!92$\% under the most extreme perturbation, whereas Gemma-4-E4B falls from $\sim\!75$\% to $\sim\!50$\% over the same sweep---a $25$-pp degradation. The InternVL3 family shows the same asymmetry: InternVL3-38B loses only $8$--$9$\,pp end-to-end, while InternVL3-8B falls from $\sim\!25$\% to $\sim\!10$\%.

Taken together, these results show that relational abstraction is sensitive to low-level visual perturbations even when the categorical identity of every object (red/blue/green; small/medium/large) is preserved. Whether this is because relations are computed over the rendered pixels or because the model's internal categorical representations become noisier remains open; internally, the same perturbation lowers the relational selectivity of late layers (\Cref{app:noise_rsa}).
See \Cref{app:noise_companion} for per-attribute differences and results on additional models.
%=============================================================
\section{Internal Mechanism: Two Competing Circuits}
\label{sec:internal}
%=============================================================

The behavioural results establish \emph{that} VLMs sit at varying positions on the object--relational axis and that several levers shift them, but not \emph{how} the choice is made internally. Our central finding is that it is not made by a single, monolithic ``relational module'' that grows stronger with scale. Rather, two distinct internal circuits\footnote{We use \emph{circuit} in a functional sense: layers and heads with a distinguishable representational signature (organising the images by object features or by relation) and a measurable causal effect on the choice under patching and ablation---not a complete, reverse-engineered computational subgraph~\citep{elhage2021mathematical, wang2022}.}---one keying on object identity, one on the abstract relation---are simultaneously active, and the model's direct answer reflects which circuit dominates. We trace the layer-wise dynamics of the two circuits (\Cref{sec:internal:competition}), show that the levers that shifted behaviour reshape them in the predicted direction (\Cref{sec:internal:n_rsa}), and confirm causally that they drive the choice (\Cref{sec:internal:cma}).
% ------------------------------------------------------------
\subsection{The two circuits compete across layer depth}
\label{sec:internal:competition}
% ------------------------------------------------------------
For every layer $\ell$, we extract the residual-stream activation at the last token of each image, yielding four hidden vectors per trial: $h_{s_1}^{(\ell)}, h_{s_2}^{(\ell)}, h_{\mathrm{rel}}^{(\ell)}, h_{\mathrm{obj}}^{(\ell)} \in \mathbb{R}^{d}$. We assemble them into a $4\!\times\!4$ image--image cosine similarity matrix~\citep{rsa, assouel2025visual}, 
$
    \mathrm{RSM}^{(\ell)}_{ij} = \cos\!\bigl(h_i^{(\ell)},\, h_j^{(\ell)}\bigr)
    \ (i, j \in \{s_1, s_2, \mathrm{rel}, \mathrm{obj}\}),
$
and compare it against two binary $4\!\times\!4$ \emph{target} matrices encoding the two competing solutions. $T_{\mathrm{rel}}$ pulls the relational match close to both samples (it instantiates the same abstract pattern); $T_{\mathrm{obj}}$ pulls the object match close to both samples (it reuses their concrete objects):
{\small
\begin{equation*}
T_{\mathrm{rel}} = 
\left(\begin{array}{c|cccc}
              & s_1 & s_2 & \mathrm{rel} & \mathrm{obj} \\
\hline
s_1           & 1   & 0   & 1            & 0   \\
s_2           & 0   & 1   & 1            & 0   \\
\mathrm{rel}  & 1   & 1   & 1            & 0   \\
\mathrm{obj}  & 0   & 0   & 0            & 1
\end{array}\right),
\qquad
T_{\mathrm{obj}} = 
\left(\begin{array}{c|cccc}
              & s_1 & s_2 & \mathrm{rel} & \mathrm{obj} \\
\hline
s_1           & 1   & 0   & 0            & 1   \\
s_2           & 0   & 1   & 0            & 1   \\
\mathrm{rel}  & 0   & 0   & 1            & 0   \\
\mathrm{obj}  & 1   & 1   & 0            & 1
\end{array}\right).
\end{equation*}}%
Note that $T_{\mathrm{rel}}$ leaves $(s_1, s_2)$ at $0$: our hypothesis concerns whether each sample sits close to the relational candidate, not how similar the two samples are to each other, and asserting the latter would reward generic image similarity unrelated to the relation.\footnote{A variant target that also clusters the two samples together yields layer-wise Rel$-$Obj curves that correlate with ours at $r=0.91$--$1.00$ for all fourteen open checkpoints (\Cref{app:rsa_target}).} We score each layer by the Pearson correlation between the upper triangle of $\mathrm{RSM}^{(\ell)}$ and that of each target:
\begin{equation*}
    \mathrm{Rel}\text{-}\mathrm{RSA}(\ell) = r\!\bigl(\mathrm{RSM}^{(\ell)}, T_{\mathrm{rel}}\bigr),
    \qquad
    \mathrm{Obj}\text{-}\mathrm{RSA}(\ell) = r\!\bigl(\mathrm{RSM}^{(\ell)}, T_{\mathrm{obj}}\bigr).
\end{equation*}
\textbf{High Rel-RSA} at layer $\ell$ means the model is organising the four images by their abstract pattern at that depth; \textbf{High Obj-RSA} means it is keying on surface features. The two scores are not redundant---a layer can score positively on neither, one, or both---and the layer-wise trajectory of $(\mathrm{Rel}\text{-}\mathrm{RSA}(\ell), \mathrm{Obj}\text{-}\mathrm{RSA}(\ell))$ reveals which solution dominates at which depth.

\begin{figure}[t]
\centering
\includegraphics[width=0.85\linewidth]{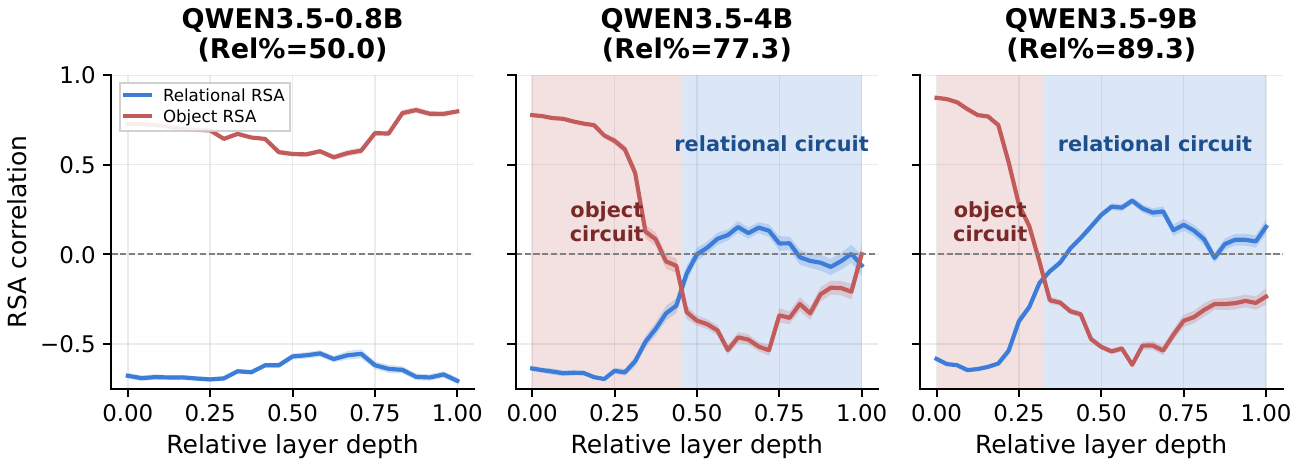}
\caption{\textbf{Layer-wise emergence of the relational circuit.} 
Per-layer $\mathrm{Rel}\text{-}\mathrm{RSA}(\ell)$ (blue) and $\mathrm{Obj}\text{-}\mathrm{RSA}(\ell)$ (red) for three Qwen3.5 checkpoints, plotted against relative layer depth. The smallest model (0.8B; Rel\%\,$=50.0$) never produces a relational circuit. The two larger models (4B, 9B; Rel\%\,$=77.3, 89.3$) exhibit an early \circbox{circuitred}{\emph{object circuit}} followed by a \circbox{circuitblue}{\emph{relational circuit}} that overtakes it at intermediate depth. The crossover shifts earlier with scale. Shaded bands: $\pm 1$ s.e.\ over trials.}
\label{fig:circuits}
\end{figure}
\autoref{fig:circuits} shows the layer-wise trajectories of $\mathrm{Rel}\text{-}\mathrm{RSA}(\ell)$ and $\mathrm{Obj}\text{-}\mathrm{RSA}(\ell)$ for three Qwen3.5 checkpoints spanning the behavioural object--relational axis: Qwen3.5-0.8B (Rel\%\,$=50.0$), Qwen3.5-4B ($77.3$), and Qwen3.5-9B ($89.3$). For Qwen3.5-0.8B, $\mathrm{Obj}\text{-}\mathrm{RSA}(\ell)$ stays high ($\sim\!0.6$--$0.8$) and $\mathrm{Rel}\text{-}\mathrm{RSA}(\ell)$ negative ($\sim\!-0.6$) at every depth, matching its chance-level Rel\%. For Qwen3.5-4B and Qwen3.5-9B, the picture is qualitatively different: $\mathrm{Obj}\text{-}\mathrm{RSA}(\ell)$ dominates the early layers (relative depth $\lesssim 0.3$--$0.4$), and $\mathrm{Rel}\text{-}\mathrm{RSA}(\ell)$ rises to overtake it at intermediate layers and remains the dominant signal through the rest of the network. We refer to these two regimes as the \circbox{circuitred}{\emph{object circuit}} (early, red shading in \autoref{fig:circuits}) and the \circbox{circuitblue}{\emph{relational circuit}} (late, blue shading). The crossover depth shifts earlier with scale ($\sim\!0.45$ in 4B, $\sim\!0.30$ in 9B), and the asymptotic Rel-RSA grows with scale, mirroring the behavioural ordering ($50.0 \to 77.3 \to 89.3$). Surface-feature matching is thus not absent in models that answer relationally; it is overwritten by the relational circuit at intermediate depth.

\begin{figure}[t]
\centering
\begin{minipage}[t]{0.38\linewidth}
    \centering
    \includegraphics[width=\linewidth]{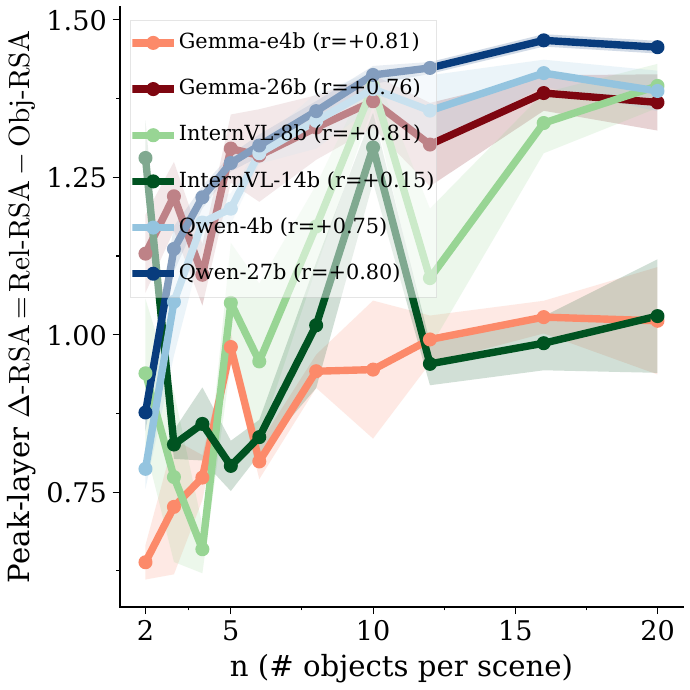}
    \caption{\textbf{The relational circuit strengthens as the number of objects per scene grows.} 
    Peak-layer gap $\Delta\text{-}\mathrm{RSA} = \max_\ell [\mathrm{Rel}\text{-}\mathrm{RSA}(\ell) - \mathrm{Obj}\text{-}\mathrm{RSA}(\ell)]$ as a function of $n$, for six open-source checkpoints. Pearson correlations with $n$ are reported in the legend. Shaded bands: $\pm 1$ s.e.\ over trials.}
    \label{fig:internal:rsa_n}
\end{minipage}\hfill
\begin{minipage}[t]{0.58\linewidth}
    \centering
    \includegraphics[width=\linewidth]{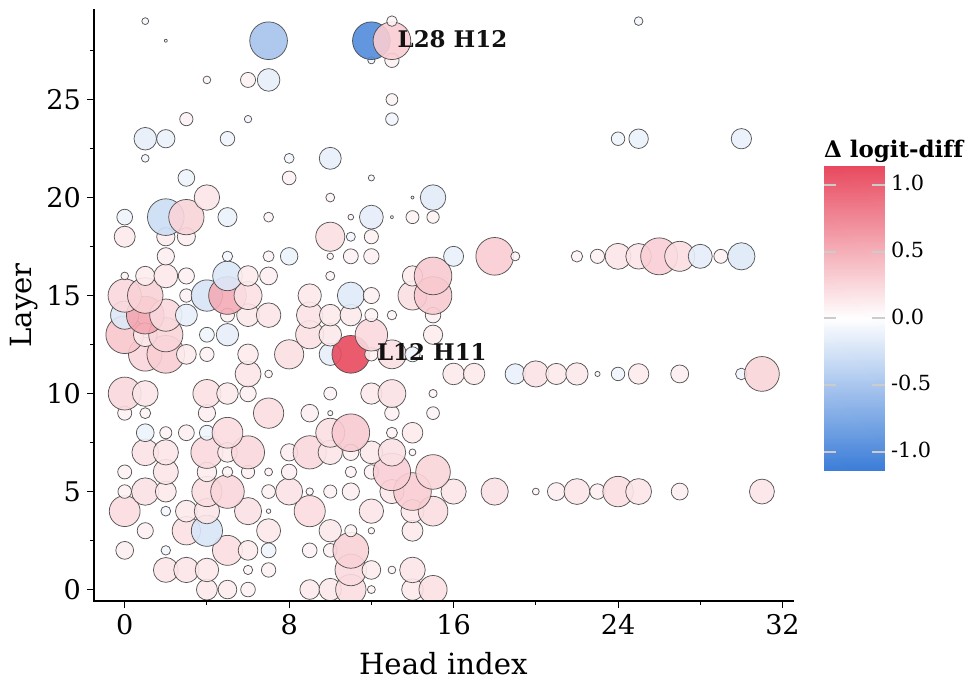}
    \caption{\textbf{Per-head activation patching effects in Gemma-4-26B.} 
    $\Delta\ell = \mathrm{logit}_A - \mathrm{logit}_B$ for each (layer, head) pair: circle size encodes $|\Delta\ell|$, colour encodes its sign (blue: relational-supporting; red: object-supporting). Two outliers stand out---\textbf{L28 H12} (deep, relational) and \textbf{L12 H11} (mid, object).}
    \label{fig:cma_head}
    \vspace{-5mm}
\end{minipage}
\end{figure}

% ------------------------------------------------------------
\subsection{The relational circuit is amplified by the number of objects}
\label{sec:internal:n_rsa}
% ------------------------------------------------------------
In \Cref{sec:behavior:n} we showed that increasing the number of objects per scene $n$ shifts the behavioral relational rate upward. We now ask whether this behavioral shift is mirrored in the internal RSA dynamics and whether the two are quantitatively linked.
\autoref{fig:internal:rsa_n} plots, for each of six open-source checkpoints, the peak-layer gap $\Delta\text{-}\mathrm{RSA} = \max_\ell \bigl[\mathrm{Rel}\text{-}\mathrm{RSA}(\ell) - \mathrm{Obj}\text{-}\mathrm{RSA}(\ell)\bigr]$ as a function of $n$.
The trend is consistent across models: $\Delta\text{-}\mathrm{RSA}$ increases with $n$, with five of the six checkpoints showing a strong positive Pearson correlation with $n$ (Gemma-4-E4B $r\!=\!+0.81$, Gemma-4-26B $r\!=\!+0.76$, InternVL3-8B $r\!=\!+0.81$, Qwen3.5-4B $r\!=\!+0.75$, Qwen3.5-27B $r\!=\!+0.80$); InternVL3-14B is the only weak case ($r\!=\!+0.15$). The lever that pushed behaviour toward the relational match thus also strengthens the relational circuit in intermediate representations, rather than merely tipping the final answer. Stimulus noise, the lever that weakened behaviour, acts on the same quantity in the opposite direction (\Cref{app:noise_rsa}).

% ------------------------------------------------------------
\subsection{Causal mediation localises the circuits to specific heads}
\label{sec:internal:cma}
% ------------------------------------------------------------

The evidence so far is correlational: $\mathrm{Rel}\text{-}\mathrm{RSA}$ and $\mathrm{Obj}\text{-}\mathrm{RSA}$ identify layers whose representations \emph{align} with the two competing solutions, but they do not establish that those layers are \emph{causally} responsible for the model's choice. To make a causal claim, we follow the mechanistic-interpretability~\citep{bereska2024mechanistic, liu2025mechanisticinterpretabilitymeets} convention of intervening on activations and measuring the resulting change in behaviour~\citep{meng2022, wang2022, assouel2025visual, heimersheim2024use, dumas-etal-2025-separating, li2026causal}.

Concretely, we run per-head activation patching on Gemma-4-26B. We form a clean context $c_1$ (relational match in position~A) and a corrupted context $c_2$ (relational match in position~B). For each (layer, head) pair $(\ell, h)$ we patch the post-$o_{\mathrm{proj}}$ activation at the prediction position from $c_2$ into $c_1$ and record the resulting shift in the logit margin $\Delta\ell = \mathrm{logit}_A - \mathrm{logit}_B$. A negative $\Delta\ell$ identifies a head that normally carries the relational signal (patching it from a corrupted run flips the model away from the relational answer); a positive $\Delta\ell$ identifies a head that carries the object signal.

\autoref{fig:cma_head} shows the resulting (layer, head) effect map. The effect is heavy-tailed: most heads have negligible $|\Delta\ell|$, while a few carry essentially all of the causal signal. The two clear outliers, \textbf{L28 H12} (deep, relational-supporting) and \textbf{L12 H11} (mid-depth, object-supporting), sit at distinct depths on opposite sides of the sign axis---consistent with the layer-wise RSA in \autoref{fig:circuits}, where the object circuit dominates early and the relational circuit takes over late. The two-stage organisation observed in representation space is thus reflected at the level of individual heads. The same organisation appears on Gemma-4-31B (\Cref{app:cma_heads}) and on five further checkpoints from Gemma-4, InternVL3 and Qwen2.5-VL, where in every case a few heads carry most of the relational effect and the strongest sits in the second half of the network (\Cref{app:cma_families}).

\paragraph{CoT bypasses the relational heads.} Zeroing the three strongest relational heads of Gemma-4-26B drops its direct answers to chance, yet with CoT the ablated model is back at $100$\% (\Cref{app:cot_bypass}): the competition governs direct answers, while the CoT gains of \Cref{sec:behavior:frontier} run through a separate, language-mediated route.

\begin{figure}[t]
\centering
\begin{minipage}[c]{0.38\linewidth}
    \centering
    \includegraphics[width=\linewidth]{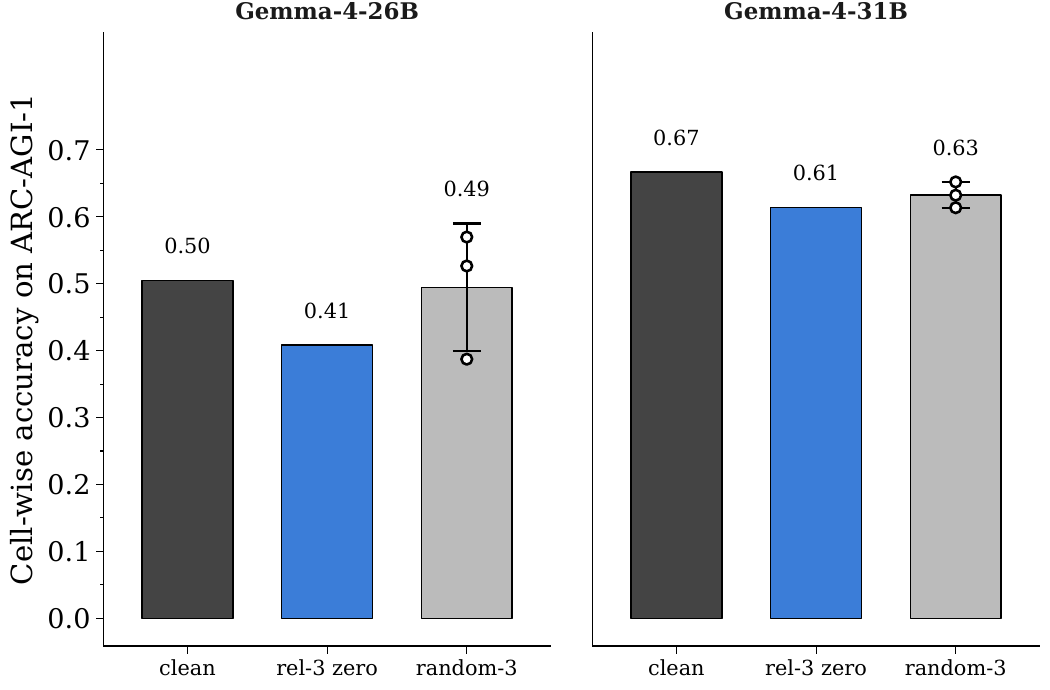}
\end{minipage}\hfill
\begin{minipage}[c]{0.58\linewidth}
    \caption{\textbf{Ablating relational heads degrades ARC-AGI-1 performance more than random ablation.} 
    Cell-wise accuracy on ARC-AGI-1 for Gemma-4-26B and -31B under \emph{clean}, \emph{rel-3 zero} (top-3 relational heads from CMA), and \emph{random-3} (three random heads) conditions. Targeted ablation drops accuracy more than random in both checkpoints. Error bars on random-3: s.d.\ across seeds.}
    \label{fig:arc_ablation}
\end{minipage}
\end{figure}

%=============================================================
\section{Discussion and Related Work}
\label{sec:discussion}
%=============================================================

\paragraph{Summary.}
Using the RMTS paradigm as a controlled probe of abstract reasoning in VLMs, \Cref{sec:behavior} showed how model capability and the strength of relational evidence in the stimulus determine where a model sits on the object--relational axis, and \Cref{sec:internal} showed that the underlying mechanism is not a single ``relational module'' but two competing circuits---one keying on surface object features, one on the abstract relation---whose relative strength is reflected layer-wise (RSA), tracked by the same levers that shifted behaviour ($n$ and noise), and causally localised to a small number of late-layer attention heads (CMA) across three model families; CoT, in contrast, reaches the relational answer without these heads. 
% The behavioural axis and the internal mechanism thus cohere into a single picture.
Extended related work is in \Cref{app:rela}.

\paragraph{Mechanistic interpretability of VLMs.}
Mechanistic interpretability~\citep{bereska2024mechanistic, sharkey2025open, zhang2026locate, nanda2023progress, olah2020zoom, elhage2021mathematical} seeks to reverse-engineer large neural networks at the level of weights, activations, and circuits, originally for LLMs and increasingly for VLMs~\citep{ben2024lvlm, liu2025mechanisticinterpretabilitymeets, liu2026seeing, fu2025hidden, yang2025ikodmitigatingvisualattention}. Recent work has localised several VLM-specific phenomena: information loss already at the vision encoder~\citep{li2025lost}, performance gains from strengthened image attention~\citep{chen2025spatialreasoninghardvlms}, and the emergence of binding mechanisms between visual tokens and concept slots~\citep{assouel2025visual}. What has remained largely uncharted is the mechanism by which a VLM extracts the abstract relation \emph{between} objects, as opposed to the identity of individual objects. Our RSA + CMA pipeline addresses this gap: we show that relation extraction is not a diffuse property of the network but is implemented by a small number of identifiable late-layer heads, sitting downstream of---and competing with---an earlier object-feature circuit. Closest to our work, \citet{lepori2024beyond} find a perceptual-then-relational decomposition of same/different judgements in vision transformers, mirroring our early object and late relational stages (\Cref{app:rela} compares the two in detail).

\paragraph{Toward AI capable of abstract reasoning.}
Recent work has repeatedly identified abstract reasoning as a core weakness of contemporary AI systems, with benchmarks such as ARC-AGI~\citep{chollet2019, chollet2025arc} highlighting the gap between human and machine performance. The gap is especially pronounced for VLMs: they underperform on ARC-style tasks~\citep{mitchell2023comparing}, fail tasks that young children solve~\citep{yiu2024kiva}, and struggle with analogical reasoning more broadly~\citep{campbell2024understanding}. By bringing the controlled RMTS paradigm to VLMs and pairing behavioural measurements with internal-state and causal analyses, we provide a more mechanistic account of \emph{which} circuits succeed and fail at relation extraction, and \emph{when}.
This account is directly actionable. As a first step, we ablated the top relational heads identified by CMA (\Cref{sec:internal:cma}) on ARC-style tasks and found that this targeted ablation degrades performance more than random head ablation of the same size (\autoref{fig:arc_ablation}; details in \Cref{app:arc_example}). The relational circuit identified on RMTS is therefore not specific to our controlled stimuli---it is recruited by the model on naturalistic abstract-reasoning benchmarks as well. We hope this kind of mechanistically informed analysis can serve as a foundation for building VLMs that genuinely abstract over objects rather than merely match their surfaces.
\subsection*{Ethics statement}
This work studies publicly released pretrained VLMs and commercial APIs on synthetic geometric stimuli; it involves no human subjects, personal data, or new data collection from people. The work is methodological---a diagnostic for relational reasoning in VLMs---and we see no direct dual-use implications. Identifying specific heads that carry the relational answer could, in principle, be used to induce or suppress particular generalisation patterns at deployment; any such edit should first be evaluated for downstream effects on unrelated tasks. Limitations are discussed in \Cref{limitation}.

\subsection*{Reproducibility statement}
The stimulus design is fully specified in \Cref{sec:task} and \Cref{app:stimuli_stats}; exact model identifiers, decoding settings, and compute are listed in \Cref{app:models}; verbatim prompts are given in \Cref{app:prompts}; the RSA and CMA procedures are defined in \Cref{sec:internal}; and the ARC-AGI-1 evaluation protocol, including the head-edit recipe, is described in \Cref{app:arc_example}; the protocols of the additional analyses are given in \Cref{app:additional}. All open-source checkpoints are publicly available, and every experiment uses greedy decoding.

\subsection*{AI use statement}
In this work, we used generative AI tools (general-purpose chat LLMs and coding assistants) to help write and edit experiment and analysis code. Additionally, we used them to draft and edit text for readability and to format the manuscript and references. Generative AI tools were not used as a component of the core methodology, and the models evaluated in this paper are the objects of study rather than research assistants. All AI-assisted code was reviewed and tested by the authors, and all AI-assisted text was checked by the authors. We take responsibility for the final content of this work, including text, claims, and artifacts produced with the aid of generative AI.

\ificlrfinal
\subsubsection*{Acknowledgments}
\textcolor{gray}{[To be added.]}
\fi

%=============================================================
% References
%=============================================================
\bibliography{refs}
\bibliographystyle{iclr2027_conference}

\appendix
%=============================================================
% APPENDIX
%=============================================================
\clearpage
\section{Trial counts and relation coverage}
\label{app:stimuli_stats}

A trial is one RMTS instance: two sample images, a relational candidate, and
an object candidate. Each image contains $n$ objects. Within an image, the
$n$ objects vary along two attributes---colour and size---and each attribute
follows one of five abstract patterns: \textsc{AAA} (all the same),
\textsc{AAB} (last different), \textsc{ABA} (outer pair shared, middle
different), \textsc{ABB} (first different), or \textsc{ABC} (all different).
A relation is therefore one pair of patterns $(P_\mathrm{color},
P_\mathrm{size})$, drawn from the resulting $5 \times 5 = 25$ relation types.

\paragraph{$n=3$.}
This is the canonical setting and the basis of every headline number in the
paper. Three positions are enough to instantiate all five patterns
unambiguously, so a single trial fixes one of the 25 relations. We then
enumerate every distinct value assignment that realises each relation---
e.g.\ \textsc{ABA} colour with $A=$ red, $B=$ blue is a different trial from
$A=$ blue, $B=$ green---and obtain \textbf{729 unique trials} in total. The
trials grade naturally by complexity: 9 of them have $P_\mathrm{color}{=}P_\mathrm{size}{=}\textsc{AAA}$
(call this $\mathrm{cpx}{=}0$, both patterns trivial), 144 have one non-trivial
attribute ($\mathrm{cpx}{=}1$), and the remaining 576 have both attributes
non-trivial ($\mathrm{cpx}{=}2$). Each trial is run twice (relational match
in position A and in position B) so the per-model evaluation totals
$729 \times 2 = 1458$ datapoints.

\paragraph{Larger $n$.}
To probe the effect of visual saliency we re-render the same 25-relation
design at $n \in \{2, 3, 4, 5, 6, 8, 10, 12, 16, 20\}$, drawing two value
assignments per relation at every $n$---i.e.\ $25 \times 2 = 50$ trials per
$n$, and 500 trials in total across the ten $n$ values. At $n{=}3$ a trial
in this sweep is structurally identical to a canonical $\mathrm{cpx}{=}2$ trial;
at $n \neq 3$ each pattern is extended deterministically (all-same to $n$
positions, ABA to ``outer pair shared, interior different,'' and so on). The
relation pool and the per-relation trial count are held fixed across $n$,
so the only thing that varies along the $n$-axis is the number of objects
per scene---any change in the relational rate therefore reflects scene
composition rather than a change in sample size.

\FloatBarrier
\section{Model details}
\label{app:models}

\paragraph{Decoding.}
All evaluations use \emph{greedy} decoding (\texttt{temperature}=0.0,
\texttt{do\_sample}=False) with \texttt{max\_tokens}=4096 (no-CoT and CoT). 
All Hugging Face IDs below are loaded with \texttt{trust\_remote\_code=True}
and run in \texttt{bfloat16} (vLLM where supported, HF Transformers
otherwise).

\begin{table}[htbp]
\centering
\small
\setlength{\tabcolsep}{6pt}
\caption{\textbf{Open-source checkpoints.} HF identifiers as released by
the model authors at the time of evaluation.}
\label{tab:hf_models}
\begin{tabular}{ll}
\toprule
Tag (used in figures) & Hugging Face identifier \\
\midrule
\multicolumn{2}{l}{\textit{Qwen3.5 (5 sizes)}} \\
qwen3.5-0.8b   & \href{https://huggingface.co/Qwen/Qwen3.5-0.8B}{Qwen/Qwen3.5-0.8B} \\
qwen3.5-2b     & \href{https://huggingface.co/Qwen/Qwen3.5-2B}{Qwen/Qwen3.5-2B} \\
qwen3.5-4b     & \href{https://huggingface.co/Qwen/Qwen3.5-4B}{Qwen/Qwen3.5-4B} \\
qwen3.5-9b     & \href{https://huggingface.co/Qwen/Qwen3.5-9B}{Qwen/Qwen3.5-9B} \\
qwen3.5-27b    & \href{https://huggingface.co/Qwen/Qwen3.5-27B}{Qwen/Qwen3.5-27B} \\
\midrule
\multicolumn{2}{l}{\textit{Gemma-4 (4 sizes)}} \\
gemma-4-e2b    & \href{https://huggingface.co/google/gemma-4-E2B-it}{google/gemma-4-E2B-it} \\
gemma-4-e4b    & \href{https://huggingface.co/google/gemma-4-E4B-it}{google/gemma-4-E4B-it} \\
gemma-4-26b    & \href{https://huggingface.co/google/gemma-4-26B-A4B-it}{google/gemma-4-26B-A4B-it} \\
gemma-4-31b    & \href{https://huggingface.co/google/gemma-4-31B-it}{google/gemma-4-31B-it} \\
\midrule
\multicolumn{2}{l}{\textit{InternVL3 (5 sizes)}} \\
internvl3-1b   & \href{https://huggingface.co/OpenGVLab/InternVL3-1B-Instruct}{OpenGVLab/InternVL3-1B-Instruct} \\
internvl3-2b   & \href{https://huggingface.co/OpenGVLab/InternVL3-2B-Instruct}{OpenGVLab/InternVL3-2B-Instruct} \\
internvl3-8b   & \href{https://huggingface.co/OpenGVLab/InternVL3-8B-Instruct}{OpenGVLab/InternVL3-8B-Instruct} \\
internvl3-14b  & \href{https://huggingface.co/OpenGVLab/InternVL3-14B-Instruct}{OpenGVLab/InternVL3-14B-Instruct} \\
internvl3-38b  & \href{https://huggingface.co/OpenGVLab/InternVL3-38B-Instruct}{OpenGVLab/InternVL3-38B-Instruct} \\
\midrule
\multicolumn{2}{l}{\textit{Qwen2.5-VL (used only in \Cref{app:cma_families})}} \\
qwen2.5-vl-7b  & \href{https://huggingface.co/Qwen/Qwen2.5-VL-7B-Instruct}{Qwen/Qwen2.5-VL-7B-Instruct} \\
\bottomrule
\end{tabular}
\end{table}

\paragraph{API checkpoints.}
The API model identifiers below are used verbatim with the official
provider SDKs.

\begin{table}[htbp]
\centering
\small
\setlength{\tabcolsep}{6pt}
\caption{\textbf{API checkpoints.}}
\label{tab:api_models}
\begin{tabular}{lll}
\toprule
Tag & Provider & Model identifier \\
\midrule
gpt-5.4               & OpenAI    & \texttt{gpt-5.4} \\
gpt-5.4-mini          & OpenAI    & \texttt{gpt-5.4-mini} \\
claude-opus-4.6       & Anthropic & \texttt{claude-opus-4-6} \\
claude-sonnet-4.6     & Anthropic & \texttt{claude-sonnet-4-6} \\
claude-haiku-4.5      & Anthropic & \texttt{claude-haiku-4-5-20251001} \\
gemini-3.1-pro        & Google    & \texttt{gemini-3.1-pro-preview} \\
gemini-3.1-flash      & Google    & \texttt{gemini-3.1-flash-lite-preview} \\
\bottomrule
\end{tabular}
\end{table}

\paragraph{Compute.}
All open-source experiments were run  with NVIDIA
H200 nodes (one GPU per job, 80--320~GB host RAM), with the runtime stack containerised in a Singularity image (PyTorch~2.4, Transformers~4.45, vLLM~0.19.1).  API-model experiments require no local GPU.

\FloatBarrier
\section{Prompt templates}
\label{app:prompts}

\paragraph{No-CoT prompt.}
The model is explicitly instructed to suppress reasoning and emit only the final answer line.

\begin{lstlisting}[basicstyle=\ttfamily\small, breaklines=true, frame=single]
I will teach you a new concept called 'XXX' using examples.

The first image is a XXX.
The second image is also a XXX.

Now look at the third image (Image A) and the fourth image (Image B).
Which one is a XXX?
The answer is equally likely to be Image A or Image B. Do NOT default to A
- examine both images carefully. Respond with ONLY one of these two lines
(no reasoning, no explanation):
Final answer: Image A
Final answer: Image B
\end{lstlisting}

\paragraph{CoT prompt.}
The model is invited to reason step by step before committing to an answer.

\begin{lstlisting}[basicstyle=\ttfamily\small, breaklines=true, frame=single]
I will teach you a new concept called 'XXX' using examples.

The first image is a XXX.
The second image is also a XXX.

Now look at the third image (Image A) and the fourth image (Image B).
Which one is a XXX?
The answer is equally likely to be Image A or Image B. Do NOT default to A
- examine both images carefully. Let's think step by step. After your
reasoning, conclude with exactly one of these two lines:
Final answer: Image A
Final answer: Image B
\end{lstlisting}

\clearpage
\section{Examples of RMTS stimuli}
\label{app:task_example}
\begin{figure}[htbp]
\centering
\includegraphics[width=0.95\linewidth]{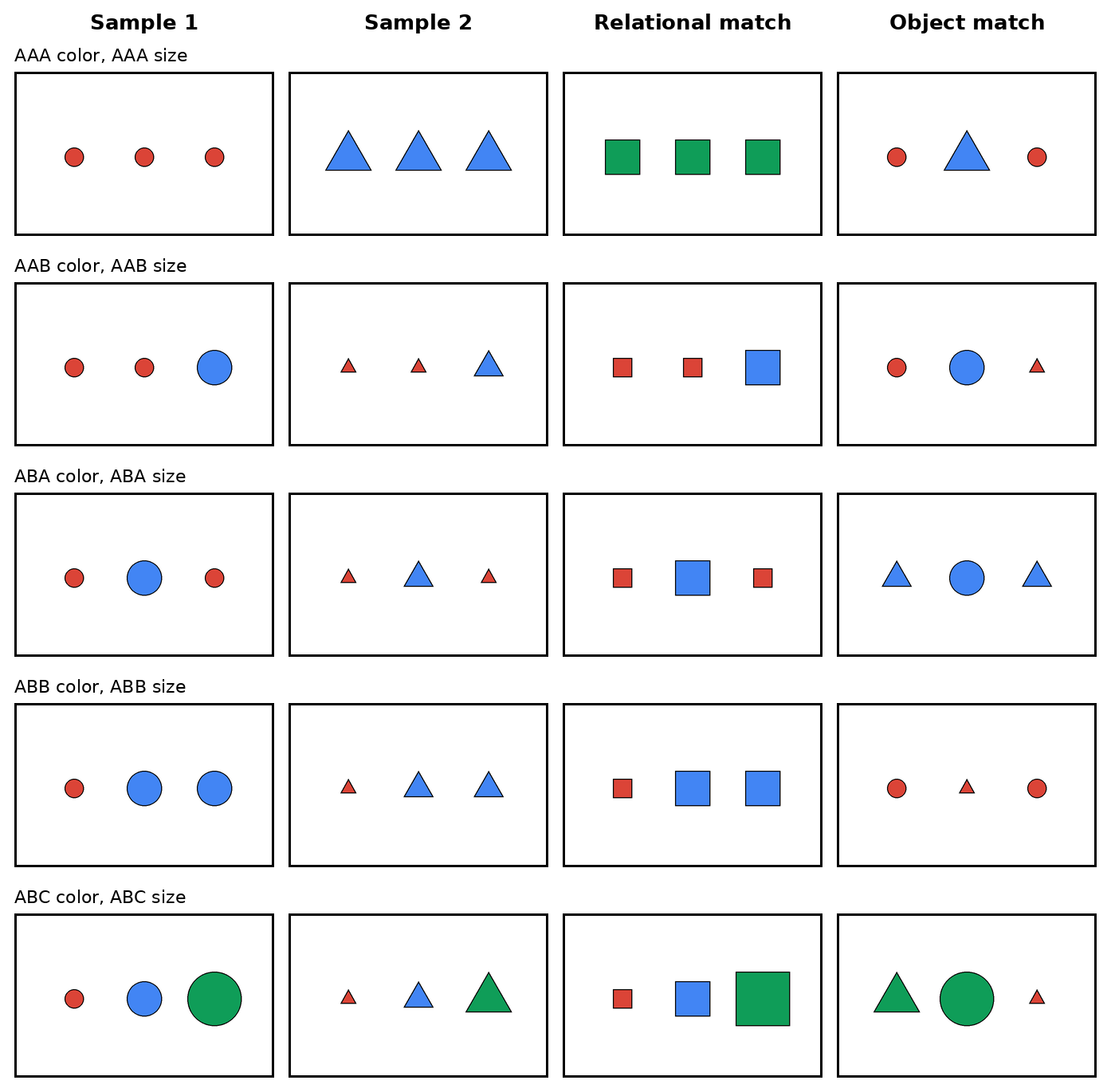}
\caption{\textbf{Canonical stimulus examples} ($n{=}3$).}
\label{fig:stimuli_canonical}
\end{figure}

\begin{figure}[htbp]
\centering
\includegraphics[width=0.95\linewidth]{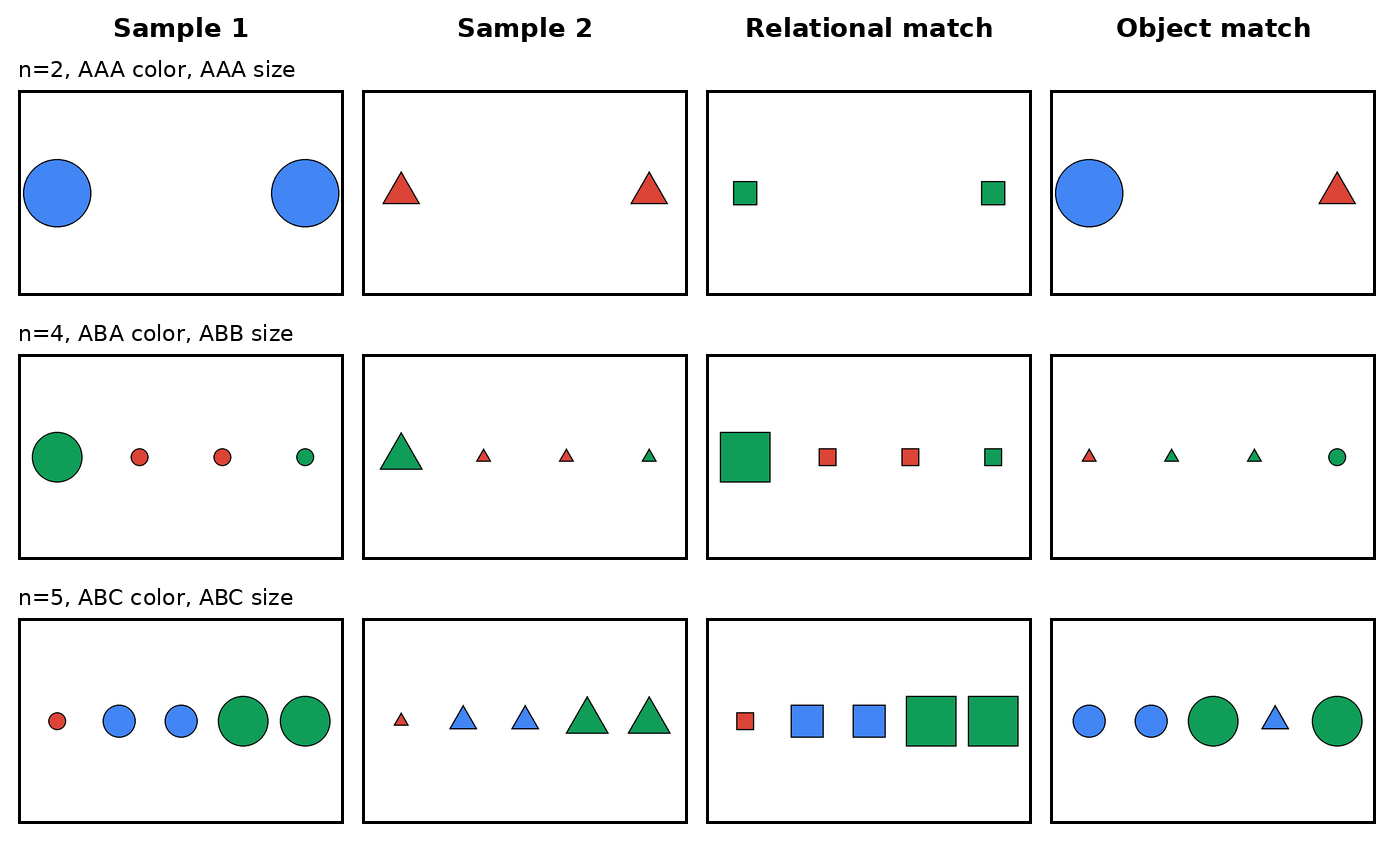}
\caption{\textbf{$n$-axis stimulus examples.} Same column layout as \autoref{fig:stimuli_canonical}, with the number of objects per scene rising from $n{=}2$ (top) to $n{=}5$ (bottom); $n{=}3$ is omitted because the canonical \autoref{fig:stimuli_canonical} already shows it.}
\label{fig:stimuli_n}
\end{figure}

\begin{figure}[htbp]
\centering
\includegraphics[width=0.96\linewidth]{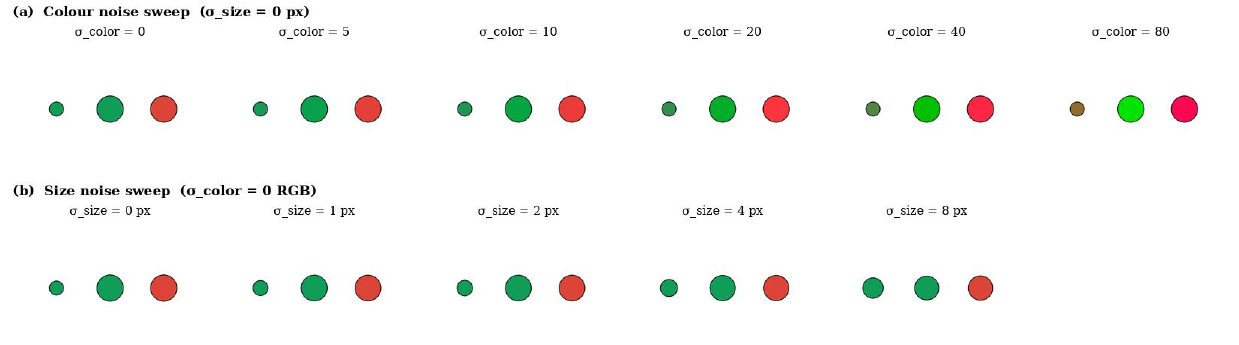}
\caption{\textbf{Per-object noise on a single cpx-2 trial (sample~1 only).}
\textbf{(a)} color noise sweep at $\sigma_s=0$: at $\sigma_c\!\geq\!40$ the categorical color identity becomes visibly ambiguous; at $\sigma_c\!=\!80$ the middle ``green'' object is closer to lime/yellow.
\textbf{(b)} size noise sweep at $\sigma_c=0$: even at $\sigma_s\!=\!8$ the small/medium/large categorical ordering survives, because the categorical sizes are 30/55/85\,px and thus 25\,px apart.
This geometric difference is the simplest explanation for the color-vs-size asymmetry reported in \Cref{sec:behavior:noise}.}
\label{fig:noise_examples}
\end{figure}

%=============================================================
\FloatBarrier
\section{Concrete CoT outputs}
\label{app:cot_outputs}
%=============================================================
\begin{figure}[htbp]
    \centering
    \includegraphics[width=1\linewidth]{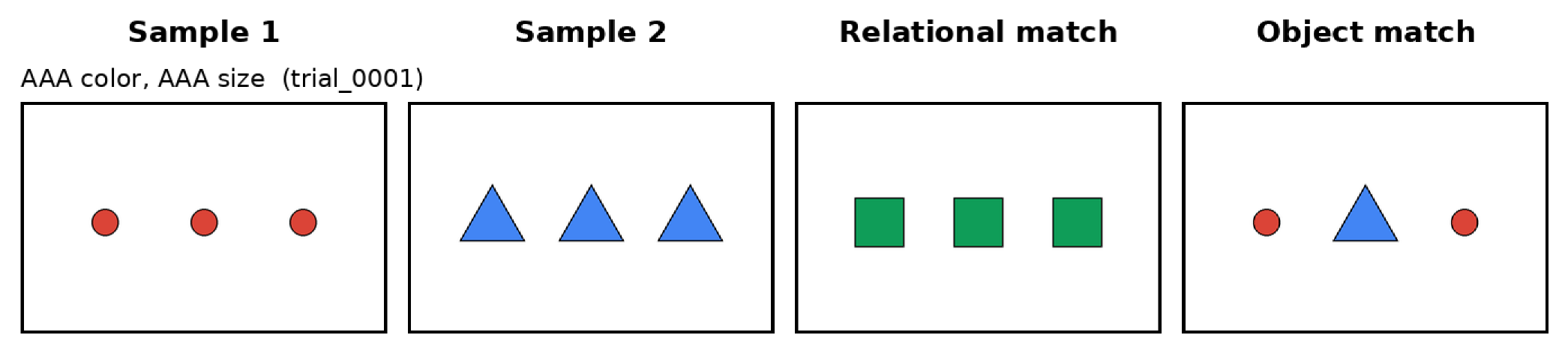}
    \caption{Example RMTS trial used to elicit the chain-of-thought trace shown below. The two top images are the samples sharing the relation $R^*$; the bottom two are the candidate test scenes (relational match and object match).}
    \label{fig:cot_example}
\end{figure}

\autoref{fig:cot_example} shows an example RMTS trial together with the corresponding chain-of-thought trace produced by Claude-Opus-4.6 on this trial. Inspecting the trace, we observe that the model explicitly extracts the pattern shared between the two sample scenes before evaluating the candidates---confirming the qualitative claim made in the main text that CoT prompting surfaces the abstract relation rather than surface-level object features.

\begin{quote}
``I need to figure out what `XXX' means from the examples. Example 1: Three red circles arranged in a horizontal row. Example 2: Three blue triangles arranged in a horizontal row. The common pattern: both examples show three identical shapes arranged in a row. The shapes are all the same\dots'' \emph{[selects relational match]}
\end{quote}

% \section{$k$ (number of in-context examples) is a weak axis}
% \label{app:k_effect}

% We varied the number of example images $k\in\{1,2,3,4\}$ for the $n$-axis study (\Cref{sec:behavior}).
% The $k$-effect is small relative to $n$: lines for different $k$ are tightly bunched within every panel of the unaggregated $n$-axis figure (\autoref{fig:behavior:n}).
% Where the effect is non-trivial, it is concentrated at the boundary $n{=}2$ in the smallest models---providing more examples weakly improves abstraction when the task is hardest.

\FloatBarrier
\section{Color-vs-size attribute asymmetry}
\label{app:attr_breakdown}

The relations in our set are defined over two attributes (color and size).
Stratifying by which attribute carries the relation (\autoref{tab:attr_breakdown}) reveals a non-trivial asymmetry under no-CoT.
Most open-source families lean \emph{color}: Gemma-4 ($+14.6$\,pp), InternVL3 ($+5.6$\,pp), Qwen3.5 ($+2.9$\,pp).
The distilled tiers from Anthropic and OpenAI lean the opposite way: Frontier (Anthropic) $-36.1$\,pp, Frontier (OpenAI distilled) $-17.4$\,pp.
The asymmetry is large enough that benchmarks dominated by a single attribute could systematically over- or under-state relational competence; we recommend evaluating on at least two orthogonal attributes whenever possible.
Chain-of-thought largely \emph{erases} this asymmetry (right half of \autoref{tab:attr_breakdown}): the Anthropic distilled tier swings from $-36.1$\,pp (no-CoT) to $0.0$\,pp under CoT, OpenAI distilled from $-17.4$ to $+4.5$, and Gemma-4 from $+14.6$ to $+0.8$.
We read this as a side-effect of the CoT lift documented in \Cref{sec:behavior}: once CoT pushes every non-saturated model close to ceiling on both attribute strata, the residual asymmetry collapses.
A practical implication: benchmarks that report CoT-only relational rates substantially understate the attribute-bias effects we identify under no-CoT.

% Combined no-CoT + CoT attribute-breakdown table.
\begin{table}[htbp]
\centering
\caption{\textbf{Color-only vs.\ size-only relational rate (\%), aggregated by family.}
``Color-only'' = trials with non-trivial color pattern but uniform size; ``size-only'' is symmetric.
For each family the rate is averaged across non-saturated members only (models at $100$\% on both attributes carry no information about the asymmetry); Frontier (Google) is at ceiling on both for every member under no-CoT, so its $\Delta$ collapses to ``---''.
Under no-CoT, families differ in the \emph{direction} of the asymmetry (Anthropic / OpenAI distilled tiers lean toward size; Gemma-4 leans toward color); under CoT every family's $|\Delta|$ collapses to $\leq 7$\,pp, indicating that explicit reasoning closes the attribute asymmetry.}
\label{tab:attr_breakdown}
\small
\setlength{\tabcolsep}{4pt}
\begin{tabular}{lrrr|rrr}
\toprule
 & \multicolumn{3}{c|}{No-CoT} & \multicolumn{3}{c}{CoT} \\
Family & Color (\%) & Size (\%) & $\Delta$ & Color (\%) & Size (\%) & $\Delta$ \\
\midrule
Frontier (OpenAI)      &  21.5 &  38.9 & $-17.4$ &  92.7 &  88.2 & $+4.5$ \\
Frontier (Anthropic)   &  63.2 &  99.3 & $-36.1$ &  98.6 &  98.6 & $+0.0$ \\
Frontier (Google)      & 100.0 & 100.0 & ---     &  83.1 &  76.4 & $+6.8$ \\
Qwen3.5                &  72.9 &  70.0 & $+2.9$  &  82.8 &  84.1 & $-1.4$ \\
Gemma-4                &  68.5 &  53.9 & $+14.6$ &  75.7 &  74.9 & $+0.8$ \\
InternVL3              &  51.0 &  45.5 & $+5.6$  &  65.3 &  66.2 & $-0.9$ \\
\bottomrule
\end{tabular}
\end{table}

\FloatBarrier
\section{$n$-axis on frontier API models (companion to \autoref{fig:behavior:n})}
\label{app:b3_api}

\begin{figure}[htbp]
\centering
\includegraphics[width=0.55\linewidth]{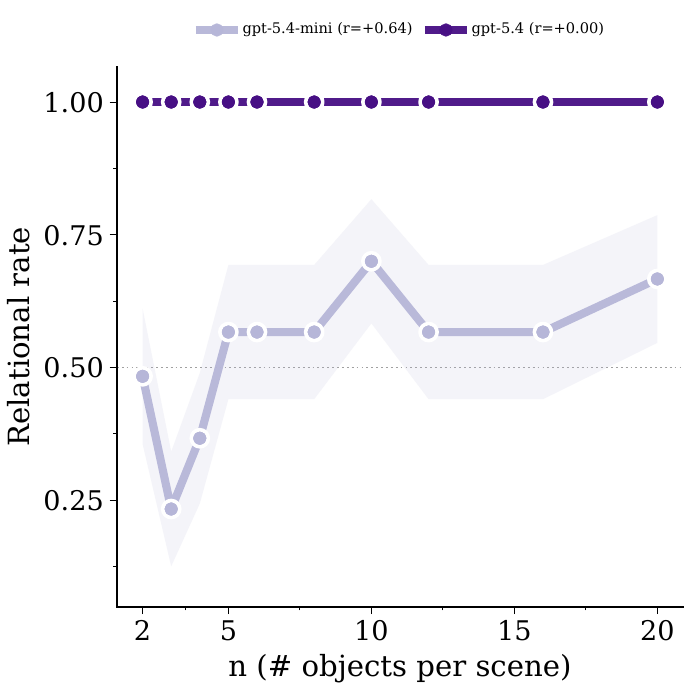}
\caption{\textbf{$n$-axis on frontier API models (no-CoT, $v2$-aligned hard $n$-axis stimuli).}
GPT-5.4 saturates at $1.0$ at every $n$. GPT-5.4-mini drops to $\approx 0.23$ at $n{=}3$---reproducing the $32$\% canonical rate up to relation-pool sampling---and rises back as $n$ grows. $95\%$ CI bands.}
\label{fig:behavior:n_api}
\end{figure}

The main-text $n$-axis figure (\autoref{fig:behavior:n}) shows the open-source families to keep the panel readable.
The corresponding curves for the frontier API models we evaluated on the $n$-axis are shown in \autoref{fig:behavior:n_api}: GPT-5.4 saturates at ceiling for every $n$, while GPT-5.4-mini exhibits the same competition the rest of the paper documents---chance at $n{=}2$ ($\approx 0.48$), object-biased at $n{=}3$ ($\approx 0.23$, matching its $32$\% rate on the canonical 729-trial set), and rising back across the chance line as $n$ grows ($\approx 0.67$ at $n{=}20$).
The two API curves and the OSS curves in \autoref{fig:behavior:n} are now evaluated on the same v2-aligned harder $n$-axis stimuli (variable shape per scene + size variation + sample-pool object match), so the $n{=}3$ point reproduces the canonical 729-trial rate up to relation-pool sampling.
The Anthropic and full Gemini families are omitted from this companion because of API-credit interruptions in the $n$-axis sweep window; the canonical 729-trial set in \Cref{sec:behavior} and the noise sweep in \Cref{sec:behavior:noise} cover them.

\FloatBarrier
\section{Shape-coherence ablation}
\label{app:shapecontrol}

\paragraph{The confound.}
The trial structure introduced in \Cref{sec:task:trial} contains a subtle shape-level asymmetry. By definition (\Cref{def:objmatch}), the object match mixes shapes from $\geq\!2$ different samples, while the relational match uses a single fresh shape. The two samples are themselves single-shape (\Cref{def:relmatch}). Concretely: the samples are shape-coherent within themselves, the relational match is shape-coherent within itself, but the object match is shape-mixed. A model could in principle exploit this asymmetry alone---``the candidate that is shape-coherent within itself, like the samples, is the answer''---and arrive at the relational match without ever inferring the colour/size relation that defines the trial.

\paragraph{The ablation.}
To test whether the relational behaviour reported in \Cref{sec:behavior} is driven by this shape-coherence cue or by genuine relation extraction, we generated an ablated stimulus set in which the object match is also rebuilt as a single-shape scene, drawn at random per trial from one of the two sample shapes. The relational match is unchanged (it still uses the brand-new third shape). Under this ablation, both candidates are shape-coherent within themselves, so shape coherence no longer disambiguates them; only the colour/size relation does. We refer to this set as \emph{shape-controlled} and to the original set used throughout \Cref{sec:behavior} as \emph{canonical}.

\paragraph{Reading the result.}
If the relational rate on the shape-controlled set matches the rate on the canonical set, the original effect cannot be driven by shape coherence. If instead the rate drops markedly, shape coherence was doing meaningful work and the relational claim should be tempered.

\paragraph{Result.}
\autoref{tab:shapecontrol} summarises the per-model relational rate on the canonical and shape-controlled sets (no-CoT). All ceiling-bound models are essentially unaffected: GPT-5.4 ($100.0 \to 100.0$), Gemini-3.1-Pro ($100.0 \to 100.0$), Gemma-4-31B ($100.0 \to 100.0$), Gemma-4-26B ($99.7 \to 99.7$, $\Delta = +0.1$\,pp), and InternVL3-38B ($100.0 \to 98.1$, $\Delta = -1.9$\,pp) reproduce their canonical-set rates within noise. Qwen3.5-9B, the only non-ceiling model in this comparison, is similarly unaffected ($89.3 \to 89.6$, $\Delta = +0.3$\,pp). All deltas are within $\pm 2$\,pp. The relational claim is therefore robust to the shape-coherence confound across the six tested checkpoints.

\begin{table}[htbp]
\centering
\small
\caption{\textbf{Shape-coherence ablation: relational rate (no-CoT) on the canonical vs.\ shape-controlled stimulus set.} On the canonical set, the object match mixes sample shapes; on the shape-controlled set, both candidates are single-shape, so shape coherence cannot disambiguate them. A small $|\Delta|$ means the relational claim is robust to the confound. All six tested checkpoints fall within $|\Delta|\!\leq\!2$\,pp.}
\label{tab:shapecontrol}
\begin{tabular}{lccc}
\toprule
Model & Canonical & Shape-controlled & $\Delta$ \\
\midrule
GPT-5.4              & 100.0 & 100.0 &  $+0.0$ \\
Gemini-3.1-Pro       & 100.0 & 100.0 &  $+0.0$ \\
Gemma-4-31B          & 100.0 & 100.0 &  $+0.0$ \\
InternVL3-38B        & 100.0 &  98.1 &  $-1.9$ \\
Gemma-4-26B          &  99.7 &  99.7 &  $+0.1$ \\
Qwen3.5-9B           &  89.3 &  89.6 &  $+0.3$ \\
\bottomrule
\end{tabular}
\end{table}

\FloatBarrier
\section{Noise sweep: additional models (API and small Qwen)}
\label{app:noise_companion}

\begin{figure}[htbp]
\centering
\includegraphics[width=0.96\linewidth]{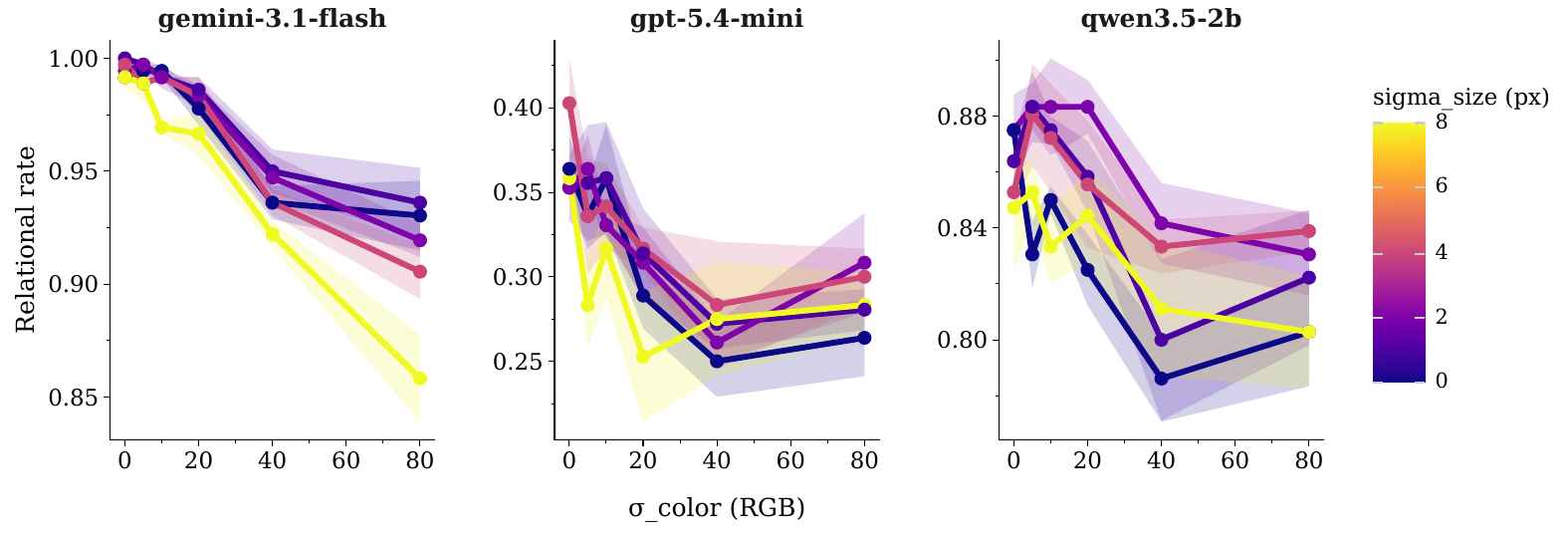}\\[2pt]
\includegraphics[width=0.96\linewidth]{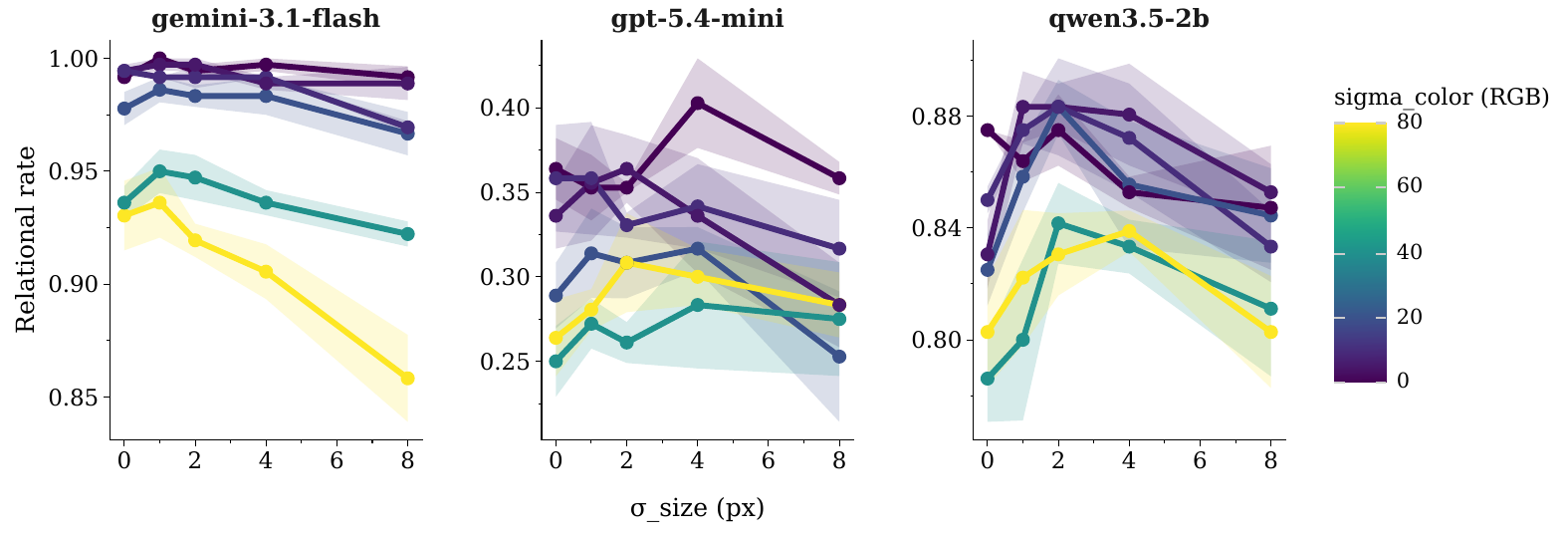}
\caption{\textbf{Noise sweep companions: API frontier models and Qwen3.5-2B.}
Top: relational rate vs.\ $\sigma_c$ (lines = $\sigma_s$). Bottom: same data plotted vs.\ $\sigma_s$ (lines = $\sigma_c$). SE bands across three seeds. Same $\sigma$-grid as \autoref{fig:behavior:noise}.}
\label{fig:behavior:noise_appendix}
\end{figure}

The main-text \autoref{fig:behavior:noise} restricts to the four open-source checkpoints that span the clean-rate range (Gemma-4-26B/E4B, InternVL3-38B/8B) so the panel reads cleanly.
For completeness we also ran the sweep on two frontier API models (Gemini-3.1-Flash, GPT-5.4-mini) and the smallest non-A-biased Qwen3.5 (2B); these companion curves are shown in \autoref{fig:behavior:noise_appendix}.
Their qualitative behaviour matches the main figure: clean rate predicts noise robustness, and color perturbation drives a substantially larger drop than size perturbation.

\paragraph{Which attribute matters.} Beyond the overall magnitude, \emph{which} attribute the noise sensitivity attaches to also varies across the models in \autoref{fig:behavior:noise}. InternVL3-38B, for instance, is almost entirely indifferent to colour noise and to size noise up to $\sigma_s\!=\!4$, only dropping appreciably at $\sigma_s\!=\!8$; its relational rate stays above $98$\% across nearly the entire sweep. Other models display the opposite asymmetry, where colour perturbations cost more than comparable size perturbations (cf.\ Gemma-4-E4B, where the $\sigma_c$ slope is sharper than the $\sigma_s$ separation).

\clearpage
\section{CMA result on Gemma-4-31B}
\label{app:cma_heads}

% \begin{figure}[htbp]
% \centering
% \begin{subfigure}[t]{0.49\linewidth}
% \includegraphics[width=\linewidth]{figures/fig_cma_layer_profile_gemma-4-26b.pdf}
% \caption{Per-layer profile, Gemma-4-26B.}
% \label{fig:cma_layer_26b_app}
% \end{subfigure}\hfill
% \begin{subfigure}[t]{0.49\linewidth}
% \includegraphics[width=\linewidth]{figures/fig_cma_layer_profile_gemma-4-31b.pdf}
% \caption{Per-layer profile, Gemma-4-31B.}
% \label{fig:cma_layer_31b_app}
% \end{subfigure}
% \caption{\textbf{Per-layer projection of the bubble plots in \autoref{fig:cma}.}
% At each layer we keep only the strongest relational head (blue, most negative $\Delta\ell$) and the strongest object head (red, most positive $\Delta\ell$). This makes the two-stage architecture (object early, rel late) more directly visible at the cost of losing the per-head spatial structure.}
% \label{fig:cma_layer_app}
% \end{figure}

\begin{figure}[htbp]
\centering
\includegraphics[width=0.72\linewidth]{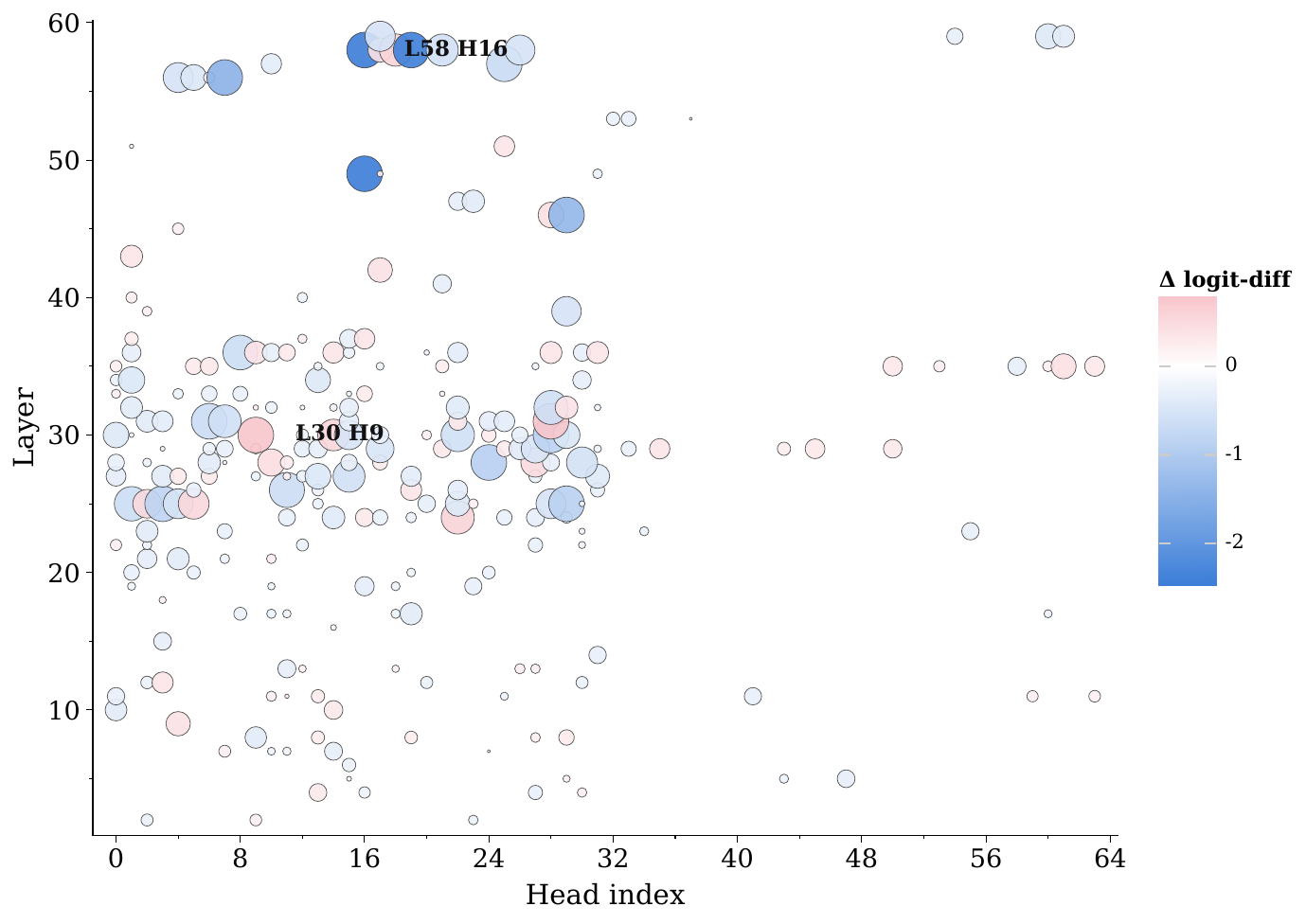}
\caption{\textbf{Per-head ablation effect on Gemma-4-31B (60 layers, 32/64 heads alternating).}
Same encoding as \autoref{fig:cma_head}: \emph{bubble size} $= |\Delta\ell|$ capped at the 95th percentile; \emph{color} $=$ signed $\Delta\ell$ (blue rel, red object).
The dominant rel head sits at \textbf{L58~H16} ($\Delta\ell = -11.7$, an order of magnitude larger than the 26B counterpart and the largest single-head effect in our experiments); dominant object heads cluster at L30--L31.
The depth at which the rel head appears scales with total depth ($\sim 95\%$ in both 26B and 31B), so the two-stage organisation is preserved across the family.
The wide gap at heads 32--63 on alternating rows again reflects Gemma-4's local/global attention pattern.}
\label{fig:cma_bubble_31b_app}
\end{figure}

% \begin{figure}[htbp]
% \centering
% \includegraphics[width=0.65\linewidth]{figures/fig_cma_topheads_gemma-4-26b.pdf}
% \caption{\textbf{Top-$10$ relational and anti-relational heads in Gemma-4-26B.} The effect is heavy-tailed: L28~H12 and L12~H11 dominate.}
% \label{fig:cma_topheads_app}
% \end{figure}

% \clearpage
% \subsection{MMMU-Pro correlation}
% \label{app:mmmu}

% We additionally correlated each open-source model's relational rate with its score on the MMMU-Pro benchmark \citep{yue2024mmmu_pro}, finding $r \in [0.84, 0.89]$ (Pearson) across the three families. Relational competence on RMTS is therefore a strong correlate of general multimodal capability, but does not reduce to it (otherwise frontier API models would not differ).

% \clearpage
\FloatBarrier
\section{Per-model relational rates}
\label{app:cot_full}

\autoref{tab:cot_full} reports the per-model no-CoT and CoT rates underlying \Cref{fig:behavior:frontier_cot,fig:behavior:scale}.
Notable rows: Claude-Haiku-4.5 ($+57.4$\,pp; $42.5\!\to\!99.9$\%) and GPT-5.4-mini ($+51.2$\,pp; $32.2\!\to\!83.4$\%) are the two largest distilled-tier gains; Gemma-4-E2B ($+57.3$\,pp) and Gemma-4-E4B ($+23.6$\,pp) show consistent within-family gains for Gemma; Qwen3.5-9B ($-4.0$\,pp), Qwen3.5-4B ($-2.4$\,pp), and InternVL3-2B ($-3.3$\,pp) show a small negative CoT effect at thin capability.

\begin{table}[htbp]
\centering
\caption{\textbf{No-CoT vs.\ CoT relational rate, full per-model breakdown.} Numbers in percent over parsed (relational$+$object) trials, excluding parse-failures and \textsc{unclear} responses; sign of $\Delta$ tells whether CoT helps (positive) or hurts (negative). Several Anthropic CoT entries are computed over the parsed subset that survived API failures during the run.}
\label{tab:cot_full}
\small
\begin{tabular}{lrrr}
\toprule
Model & no-CoT (\%) & CoT (\%) & $\Delta$ \\
\midrule
GPT-5.4               & 100.0 &  99.8 & $-0.2$ \\
GPT-5.4-mini          &  32.2 &  83.4 & $+51.2$ \\
Claude-Opus-4.6       & 100.0 & 100.0 & $+0.0$ \\
Claude-Sonnet-4.6     & 100.0 & 100.0 & $+0.0$ \\
Claude-Haiku-4.5      &  42.5 &  99.9 & $+57.4$ \\
Gemini-3.1-Pro        & 100.0 & 100.0 & $+0.0$ \\
Gemini-3.1-Flash      &  99.7 &  99.9 & $+0.2$ \\
\midrule
Qwen3.5-0.8B          &  50.0 &  49.5 & $-0.5$ \\
Qwen3.5-2B            &  47.4 &  52.2 & $+4.8$ \\
Qwen3.5-4B            &  77.3 &  74.9 & $-2.4$ \\
Qwen3.5-9B            &  89.3 &  85.3 & $-4.0$ \\
Qwen3.5-27B           &  85.9 &  86.6 & $+0.7$ \\
\midrule
Gemma-4-E2B           &  26.3 &  83.6 & $+57.3$ \\
Gemma-4-E4B           &  74.3 &  97.9 & $+23.6$ \\
Gemma-4-26B           &  99.7 & 100.0 & $+0.3$ \\
Gemma-4-31B           & 100.0 & 100.0 & $+0.0$ \\
\midrule
InternVL3-1B          &  50.1 &  48.5 & $-1.6$ \\
InternVL3-2B          &  49.5 &  46.1 & $-3.3$ \\
InternVL3-8B          &  28.9 &  51.7 & $+22.8$ \\
InternVL3-14B         &  54.1 &  90.6 & $+36.5$ \\
InternVL3-38B         & 100.0 &  98.8 & $-1.2$ \\
\bottomrule
\end{tabular}
\end{table}

\FloatBarrier
\section{ARC-AGI-1 task detail}
\label{app:arc_example}

\begin{figure}[htbp]
\centering
\includegraphics[width=0.95\linewidth]{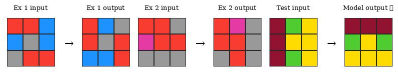}
\caption{\textbf{One ARC-AGI-1 training task as the model sees it.}
Two of the four train pairs (input $\to$ output) and the test input are shown to the model as rendered colour images; the rule here is grid \emph{transposition} (rows become columns).}
\label{fig:arc_example}
\end{figure}

\autoref{fig:arc_ablation} tests whether the
relational heads we identified on RMTS make a measurable contribution on a fully out-of-domain abstract-reasoning benchmark.  This appendix gives the exact stimulus, prompt, evaluator, and head-edit recipe.

\subsection{Dataset and task subset}

We use the public \textbf{ARC-AGI-1} corpus~\citep{chollet2019}
(\texttt{training} split; $400$ tasks).  Each task contains $2$--$5$ train
input/output pairs and one test input; the model must emit the test output
grid.  Two filters are applied before evaluation:
\begin{itemize}
\item \textbf{Image budget}: the prompt contains $2K{+}1$ images for $K$
      train pairs.  We cap at $14$ images per prompt
      (\texttt{--limit-mm-per-prompt 14}), which covers $K{\le}6$ pairs.
\item \textbf{Task cap}: after the image filter we take the first
      \texttt{LIMIT} tasks in canonical filename order
      ($66$ for Gemma-4-26B, $80$ for Gemma-4-31B; the difference is purely
      a job-budget choice and has no effect on the comparison since each
      condition uses the same $n$ within a model).
\end{itemize}
Identical task subsets are evaluated under every head-edit condition for a
given model, so all condition deltas are paired.

\subsection{Stimulus rendering}

Each $H{\times}W$ grid is rendered as an RGB image with one $32{\times}32\,$px
cell per integer ($0$--$9$) using the canonical ARC palette
($0\!=\!$black, $1\!=\!$blue, $2\!=\!$red, $3\!=\!$green, $4\!=\!$yellow,
$5\!=\!$grey, $6\!=\!$pink, $7\!=\!$orange, $8\!=\!$light-blue,
$9\!=\!$maroon), with a $1\,$px dark border around each cell.  The model never
sees the integer matrix---it sees the rendered colour image, the same modality
used throughout the rest of the paper.

\subsection{Prompt format}

A single user turn carries an instruction header, the palette key, and an
interleaved sequence of (image, shape annotation) blocks for each train pair
followed by the test input.  Verbatim:

\begin{quote}\small
You are given input/output examples that follow a hidden rule.  Each example
is shown as a coloured grid; each cell is one integer 0--9.  \emph{[palette
line]}.  Infer the rule from the examples and apply it to the final test
input.  Be concise and decisive: do NOT second-guess yourself or restate the
examples.  After your reasoning, output ONLY the final grid for the test
input between the literal markers \texttt{<grid>} and \texttt{</grid>}: one
row per line, integers 0--9 separated by spaces, no row numbers, no colour
names, no extra text. \\
Example 1 input: \emph{[image]} \quad ($H_1{\times}W_1$ grid) \\
Example 1 output: \emph{[image]} \quad ($H'_1{\times}W'_1$ grid) \\
\ldots \\
Test input: \emph{[image]} \quad ($H_t{\times}W_t$ grid). Produce the test
output grid \ldots between \texttt{<grid>}\ldots\texttt{</grid>}.
\end{quote}

We do \emph{not} use chain-of-thought (\texttt{--cot} flag off) for the
intervention experiments, because (a)~the rel-3 ablation we want to detect is
a \emph{representational} effect that should not require verbal scratchpad to
manifest, and (b)~CoT lengthens the response, increasing the chance that any
head edit becomes visible only via reasoning errors rather than directly via
the predicted grid.

\subsection{Decoding and parsing}

We run the unquantised checkpoint via HuggingFace Transformers (no vLLM in
the intervention path, since vLLM does not expose pre-\texttt{o\_proj}
hooks).  Decoding is greedy (\texttt{do\_sample=False}, no temperature) with
\texttt{max\_new\_tokens=4096}.  The parser:
\begin{itemize}
\item First tries the explicit \texttt{<grid>}\ldots\texttt{</grid>} block.
\item Falls back to the last \texttt{Output:} / \texttt{Answer:} block.
\item For each line, extracts integer tokens; if none are found, tries the
      colour-name fallback (e.g., ``black blue red'' $\to$ $[0,1,2]$,
      including the two-word phrases ``light blue'' and ``dark red'').
\end{itemize}

\subsection{Metric: cell-wise accuracy}

ARC-AGI-1's official metric is exact-match (whole grid correct or incorrect),
but exact-match is too lossy at the capability tier of Gemma-4-26B
(clean exact-match $\approx 0.23$): every condition crashes to near zero and
the rel-vs-random comparison becomes statistically uninformative.  We
therefore report \textbf{cell-wise accuracy}---the fraction of cells in the
ground-truth grid matched by the parsed grid, counted as $0$ if the parsed
shape does not match.  This is a natural relaxation that preserves ordering
on hard tasks (a one-cell-wrong answer scores $\frac{HW{-}1}{HW}$, not $0$)
while still penalising shape errors.

\subsection{Head-edit recipe}

We use the same per-head intervention machinery as our RMTS CMA scan
(\Cref{sec:internal:cma}): a forward-pre-hook on each layer's
\texttt{o\_proj} reshapes its input to $(B,T,n_h,d_h)$ and multiplies the
selected head slots by the chosen scale.  In every condition reported in the
main text the scale is $0$ (full ablation).  
% The (layer, head) sets, taken
% from the top of the bubble plot in \autoref{fig:cma}:

\begin{itemize}
\item \textbf{Gemma-4-26B (30 layers, 16/32 heads)}: \emph{rel-3 zero}:
      L28 H12, L28 H7, L19 H2.  \emph{obj-3 zero}: L12 H11, L14 H1, L15 H5.
      \emph{Random-3} (three seeds): each is three (layer, head) pairs sampled
      reproducibly from the layer-by-head grid excluding the top-5 rel and
      top-5 object heads, so each random condition zeros out the same number
      of head slots as the rel/obj conditions.
\item \textbf{Gemma-4-31B (60 layers, 32/64 heads)}: \emph{rel-3 zero}:
      L58 H16, L58 H19, L49 H16 (L58~H16 has $\Delta\ell=-11.7$ in the CMA
      scan, four-to-five times the runner-up).  \emph{obj-3 zero}: L30 H9,
      L30 H14, L31 H28.  Random-3 sampled analogously.
\end{itemize}

The random-3 baseline controls for the trivial ``editing any head hurts''
explanation: if rel-3 ablation drops cell-wise accuracy beyond what
random-3 ablation does, the rel circuit is contributing causal capacity to
ARC, not just disturbing the residual stream.

\FloatBarrier
\section{Additional analyses}
\label{app:additional}

This appendix collects analyses that probe the generality of the behavioural and mechanistic findings: a structurally different relation (\Cref{app:numeq}), causal mediation on further model families (\Cref{app:cma_families}), the interaction between CoT and the relational heads (\Cref{app:cot_bypass}), the internal effect of stimulus noise (\Cref{app:noise_rsa}), and the robustness of the RSA target (\Cref{app:rsa_target}).

\subsection{A structurally different relation: numerical equality}
\label{app:numeq}

\paragraph{Design.} All relations in the main text are ordinal patterns over colour and size. To test whether the findings depend on this relation type, we construct trials in which the two samples share their \emph{object count} and nothing else. With three objects, for example, Sample~1 shows three red squares and Sample~2 three blue triangles, so that count is their only commonality. The relational match shows three green circles---the right count, in a colour and shape used by neither sample---while the object match shows five red squares, Sample~1's objects copied verbatim with the wrong count. A model that matches appearance picks the red squares; only counting reaches the green circles. As in the main experiments, every trial is presented in both answer orderings, so any positional preference cancels (chance $=50$\%), and no CoT is used.

\paragraph{Scale effect on open-source models.} We generate $64$ trials ($128$ judgements per model) and evaluate two checkpoints from each open-source family (\autoref{tab:numeq_oss}). In every family the relational rate rises with model size, by $46$, $30$ and $73$\,pp respectively, so the scale effect of \Cref{sec:behavior:scale} is robust to changing the relation itself.

\paragraph{Frontier models come off the ceiling.} The three frontier models that saturate the ordinal relations are no longer saturated on numerical equality (\autoref{tab:numeq_api}). Together with the noise sweep, in which checkpoints that saturate the clean stimuli also fall off the ceiling (Gemma-4-26B $100\to92$\%, InternVL3-38B $100\to92$\%, Gemini-3.1-Flash $99\to86$\% under the strongest colour and size jitter; \autoref{fig:behavior:noise} and \Cref{app:noise_companion}), this indicates that the ceiling on the canonical set partly reflects its simplicity for the strongest models, and that harder relations discriminate among them.

\begin{table}[htbp]
\centering
\small
\begin{minipage}[t]{0.5\linewidth}
\centering
\caption{\textbf{Numerical equality, open-source models.} Relational rate (\%, no-CoT), small $\to$ large checkpoint per family.}
\label{tab:numeq_oss}
\begin{tabular}{lcc}
\toprule
Family (small $\to$ large) & Small & Large \\
\midrule
Qwen3.5 (2B $\to$ 9B)      & 47.7 & 93.8 \\
Gemma-4 (E4B $\to$ 26B)    & 26.6 & 57.0 \\
InternVL3 (8B $\to$ 38B)   & 25.8 & 98.4 \\
\bottomrule
\end{tabular}
\end{minipage}\hfill
\begin{minipage}[t]{0.44\linewidth}
\centering
\caption{\textbf{Numerical equality, frontier models.} Relational rate (\%), averaged over $n=2$--$5$ and both orderings.}
\label{tab:numeq_api}
\begin{tabular}{lc}
\toprule
Model & Rel.\ rate \\
\midrule
Gemini-3.1-Pro   & 99.0 \\
Claude-Opus-4.6  & 93.0 \\
GPT-5.4          & 82.5 \\
\bottomrule
\end{tabular}
\end{minipage}
\end{table}

\subsection{Causal mediation on further model families}
\label{app:cma_families}

To test whether the head-level picture of \Cref{sec:internal:cma} is specific to Gemma-4-26B, we run the same per-head activation patching, unchanged, on five further checkpoints covering Gemma-4, InternVL3 and Qwen2.5-VL ($30$ trials each, both answer orderings). \autoref{tab:cma_families} reports the strongest relational head per checkpoint. In every family a small number of heads carry most of the relational effect, and in every case the strongest one sits in the second half of the network, as in the Gemma-4-26B result of the main text.

\begin{table}[htbp]
\centering
\small
\caption{\textbf{Strongest relational head per checkpoint.} Effect: change in the log-odds of the relational answer when that head alone is patched (negative = relational-supporting, as in \autoref{fig:cma_head}). Depth: layer index divided by the total number of layers.}
\label{tab:cma_families}
\begin{tabular}{lccc}
\toprule
Model & Top relational (layer, head) & Effect & Depth \\
\midrule
Gemma-4-26B (main text) & (28, 12) & $-1.02$ & 0.97 \\
Gemma-4-E4B             & (21, 5)  & $-1.27$ & 0.51 \\
InternVL3-38B           & (55, 23) & $-1.36$ & 0.87 \\
InternVL3-14B           & (28, 29) & $-0.82$ & 0.60 \\
InternVL3-8B            & (17, 27) & $-0.44$ & 0.63 \\
Qwen2.5-VL-7B           & (20, 24) & $-0.32$ & 0.74 \\
\bottomrule
\end{tabular}
\end{table}

\subsection{CoT bypasses the relational heads}
\label{app:cot_bypass}

CoT lifts weaker models toward the relational match (\Cref{sec:behavior:frontier}). Does it do so by strengthening the relational circuit, or through a different route? On Gemma-4-26B we zero out the relational heads identified by CMA (the single strongest, L28~H12, or the top three, L28~H12, L28~H7 and L19~H2; \Cref{app:arc_example}) and run the identical trials twice, once answering directly and once with the CoT prompt of \Cref{app:prompts}. We use $30$ trials from the most demanding relation subset in both answer orderings ($60$ judgements per condition). \autoref{tab:cot_bypass} shows that removing the three strongest relational heads reduces direct answers to chance, which also shows that these heads are necessary rather than merely correlated with the behaviour. With CoT, however, the same ablated model is back at $100$\%. The CoT gain therefore does not run through the circuit we identified.

\begin{table}[htbp]
\centering
\small
\caption{\textbf{Relational rate (\%) of Gemma-4-26B under head ablation, without and with CoT.}}
\label{tab:cot_bypass}
\begin{tabular}{lcc}
\toprule
Heads zeroed & Direct answer & With CoT \\
\midrule
None (intact model)                 & 100.0 & 100.0 \\
Strongest relational head           &  86.7 & 100.0 \\
Three strongest relational heads    &  50.0 & 100.0 \\
\bottomrule
\end{tabular}
\end{table}

\subsection{Stimulus noise and internal representations}
\label{app:noise_rsa}

We ask whether stimulus noise, which lowers the relational rate (\Cref{sec:behavior:noise}), acts on the same internal quantity as the number of objects (\Cref{sec:internal:n_rsa}). We use Gemma-4-E4B, the checkpoint with the largest behavioural noise effect, and apply independent Gaussian jitter to every object's radius, keeping each object inside its original size category so that the categorical description of the scene is unchanged and only the rendered pixels move. We use three noise levels, from no jitter to the strongest jitter of the behavioural sweep, with three noise seeds and $40$ trials per seed ($120$ trials per condition), and measure the relational selectivity $\Delta\text{-}\mathrm{RSA} = \mathrm{Rel}\text{-}\mathrm{RSA} - \mathrm{Obj}\text{-}\mathrm{RSA}$ used throughout \Cref{sec:internal}. Averaged over the late third of the layers (layers 29--42), $\Delta\text{-}\mathrm{RSA}$ falls from $+0.24$ without jitter to $+0.15$ at the strongest jitter (s.e.\ $0.06$), while the relational rate on the same trials falls from $78$\% to $69$\%. Noise thus acts on the same internal quantity as $n$, in the opposite direction, and the internal decline tracks the behavioural one.

\subsection{Robustness of the relational RSA target}
\label{app:rsa_target}

The relational target $T_{\mathrm{rel}}$ (\Cref{sec:internal:competition}) leaves the sample--sample entry at $0$. We also evaluate the alternative in which all relation-sharing images, including the two samples, are clustered together. For each model and each target we compute the layer-wise curve $\mathrm{Rel}\text{-}\mathrm{RSA}(\ell) - \mathrm{Obj}\text{-}\mathrm{RSA}(\ell)$ and take the Pearson correlation across layers between the two curves. The two definitions agree at $r=0.91$--$1.00$ for all fourteen open checkpoints across the three families, so every layer-wise conclusion in the paper holds under either target.

\FloatBarrier
\section{Limitations}
\label{limitation}
\paragraph{Shape coherence as a residual cue.} Within a trial, every sample scene uses one shape, the relational match introduces a fresh shape, and the object match by construction (\autoref{def:objmatch}) mixes $\geq\!2$ sample shapes; this enforces shape-disjointness with the samples but also means the relational candidate is shape-coherent while the object candidate is shape-mixed. A model could therefore exploit a ``which candidate is shape-coherent like the samples'' heuristic without inferring the colour/size relation. We address this with a candidate-level ablation in \Cref{app:shapecontrol}: the object match is rebuilt as single-shape (drawn from a sample shape) so both candidates are shape-coherent within themselves, and only the colour/size relation can disambiguate them. Across the six tested checkpoints the relational rate moves by at most $|\Delta|\!\leq\!2$\,pp, indicating that the headline effect is not driven by shape coherence.
\paragraph{Stimulus distribution.} Our formal geometric stimuli isolate the relational step at the cost of a narrow distribution; we mitigate this by sweeping $n\!=\!2$--$20$ across the full $5\!\times\!5$ pattern grid (\autoref{fig:behavior:n}), by adding per-object Gaussian noise (\autoref{fig:behavior:noise}), and by testing a structurally different relation, numerical equality (\Cref{app:numeq}). These relations still do not exhaust those VLMs will ultimately need; spatial, causal and semantic relations, as well as naturalistic scenes, are an immediate next step.
\paragraph{Ceiling effects.} The strongest frontier models are at ceiling on the ordinal colour/size relations. For these closed-source models the stimulus set functions only as a benchmark, and at ceiling it cannot discriminate genuine relational abstraction from simpler strategies; the numerical-equality relation partly addresses this by bringing them off the ceiling (\Cref{app:numeq}).
\paragraph{Mechanistic scope.} Per-head CMA evidence covers two Gemma-4 checkpoints in full (\Cref{sec:internal:cma,app:cma_heads}) and five further checkpoints from Gemma-4, InternVL3 and Qwen2.5-VL on a smaller trial budget (\Cref{app:cma_families}). Our use of \emph{circuit} is functional (\Cref{sec:internal}); we do not characterise how the identified heads compose. Finally, the circuit account covers direct answers only: under CoT the model reaches the relational answer without the relational heads (\Cref{app:cot_bypass}), and the mechanism of that route is left open. 
% the patching code is family-specific---Gemma-4's local/global attention pattern requires custom hook plumbing---and a non-Gemma replication on InternVL3 is in progress. The current patching contrast (relational match in position A vs.\ B) does not fully separate ``carries the relational signal'' from ``carries answer-position information''; counterfactuals that hold position fixed while varying relational content would tighten the causal interpretation.

\FloatBarrier
\section{Extended Related Work}
\label{app:rela}
\paragraph{Analogical reasoning.}
The RMTS task is treated in cognitive science as an instance of \emph{analogical reasoning}~\citep{LanguageHelpsChildren, analogyandabstraction}, in which a learner transfers an abstract structural correspondence from a source to a target despite surface-level dissimilarity. Analogy has been studied as a candidate explanation for the remarkable sample efficiency of human learning~\citep{Thagard1992AnalogyEducation, analogyandabstraction} and for its role in scientific discovery~\citep{Leatherdale1974, analogycreative1997goel, analogyandcreative}.

Studies of analogical reasoning in deep neural networks remain comparatively scarce, but several lines of work are directly relevant. \citet{webb2023emergent} compare large language models to humans on a battery of analogical-reasoning tasks and report broadly human-level behavioural performance. \citet{campbell2024understanding} document VLM failures on visual analogy benchmarks, in line with the broader picture that vision-grounded analogical inference remains a weakness of current systems. From a more mechanistic angle, \citet{minegishi2025analogy} formalise analogy in a category-theoretic framework, design a controlled toy task, and analyse how a Transformer trained from scratch on this task implements analogical mappings.

The closest mechanistic study is \citet{lepori2024beyond}, who show that vision transformers decompose a same/different judgement into a perceptual stage followed by a relational stage, close to the early object stage and late relational stage we describe. We differ in three respects: we study generative VLMs on a task where the relational answer must beat a specific object-matching alternative rather than give a binary report; we track that competition across capability tier, scale, CoT, scene size and stimulus noise; and we test the relational stage causally, including on ARC-AGI-1.

These prior efforts are largely either purely behavioural (assessing whether a model gets analogy problems right or wrong) or focused on toy networks trained from scratch on synthetic tasks. To our knowledge, the present work is the first to combine \textbf{(i)} a parametrically controlled analogical-reasoning paradigm imported from comparative and developmental psychology, \textbf{(ii)} an evaluation across the current frontier of pretrained VLMs (API and open-source families), and \textbf{(iii)} a mechanistic dissection that traces how the analogical computation is implemented inside the network at the level of layers and individual attention heads.
\paragraph{Relational inductive biases and architectures.}
A complementary line of work builds architectures with explicit relational inductive biases, aiming for systematic generalisation beyond surface features. \citet{webb2023ocra} propose Object-Centric Relational Abstraction (OCRA), combining slot-based object representations with an explicit relational module for abstract visual reasoning. \citet{webb2024relationalbottleneck} synthesise and extend this theme via the \emph{relational bottleneck}, an architectural constraint that encourages abstraction by forcing computation to pass through a relation-focused representation. \citet{mondal2024slotabstractors} scale these ideas to more complex multi-object settings by combining slot-based perception with Transformer-style abstraction.

\paragraph{Competing internal circuits.}
The two-circuit picture we observe is reminiscent of similar phenomena
documented elsewhere in the mechanistic-interpretability literature.
Most strikingly, the grokking phenomenon~\citep{liu2023omnigrok, power2022grokkinggeneralizationoverfittingsmall} has been explained as a
competition between a memorisation circuit and a generalisation
circuit, with the latter eventually dominating after extended
training~\citep{nanda2023progress, liu2022towards, minegishi2023bridging, merrill2023tale}. Our findings
suggest that the same competitive logic surfaces at inference time in
pretrained VLMs: a surface-feature (object) circuit and an abstract
(relational) circuit are both present, and the model's behavioural
output reflects which circuit dominates at the prediction position.
We conjecture that large pretrained models harbour many such
circuits implementing alternative solutions to the same task,
and that the model's response is governed by whichever circuit
exerts the largest causal influence on the output logits.

\end{document}